\documentclass{article}

\usepackage{microtype}
\usepackage{graphicx}
\usepackage{subcaption}
\usepackage{cuted}
\usepackage{placeins}
\usepackage{booktabs} % for professional tables
\usepackage[normalem]{ulem}

\usepackage{hyperref}
\usepackage{booktabs}
\usepackage{tabularx}
\usepackage[most]{tcolorbox}

\usepackage[preprint]{icml2026}

\usepackage{amsmath}
\usepackage{amssymb}
\usepackage{mathtools}
\usepackage{amsthm}
\usepackage{enumitem}
\setlist{itemsep=0pt, topsep=2pt, parsep=0pt}

\usepackage[capitalize,noabbrev]{cleveref}

\usepackage[textsize=tiny]{todonotes}

\usepackage{tcolorbox}
\usepackage{amsthm}
\usepackage{amssymb}
\usepackage{csquotes}

\newtcolorbox{theorembox}{
  colback=gray!10,
  colframe=gray!40,
  boxrule=0.5pt,
  arc=1pt,
  left=8pt, right=8pt, top=6pt, bottom=6pt
}

\newtheorem{theorem}{Theorem}

\icmltitlerunning{Modern GLOM: Canonical Locks that encode part-whole hierarchies}

\newsavebox{\teaserbox}

\begin{document}

\sbox{\teaserbox}{%
  \begin{tcolorbox}[enhanced, width=0.97\textwidth,
      colback=white, colframe=black!12, boxrule=0.5pt, arc=4pt,
      boxsep=0pt, left=7pt, right=7pt, top=7pt, bottom=7pt,
      drop fuzzy shadow=black!10]
    \includegraphics[width=\linewidth]{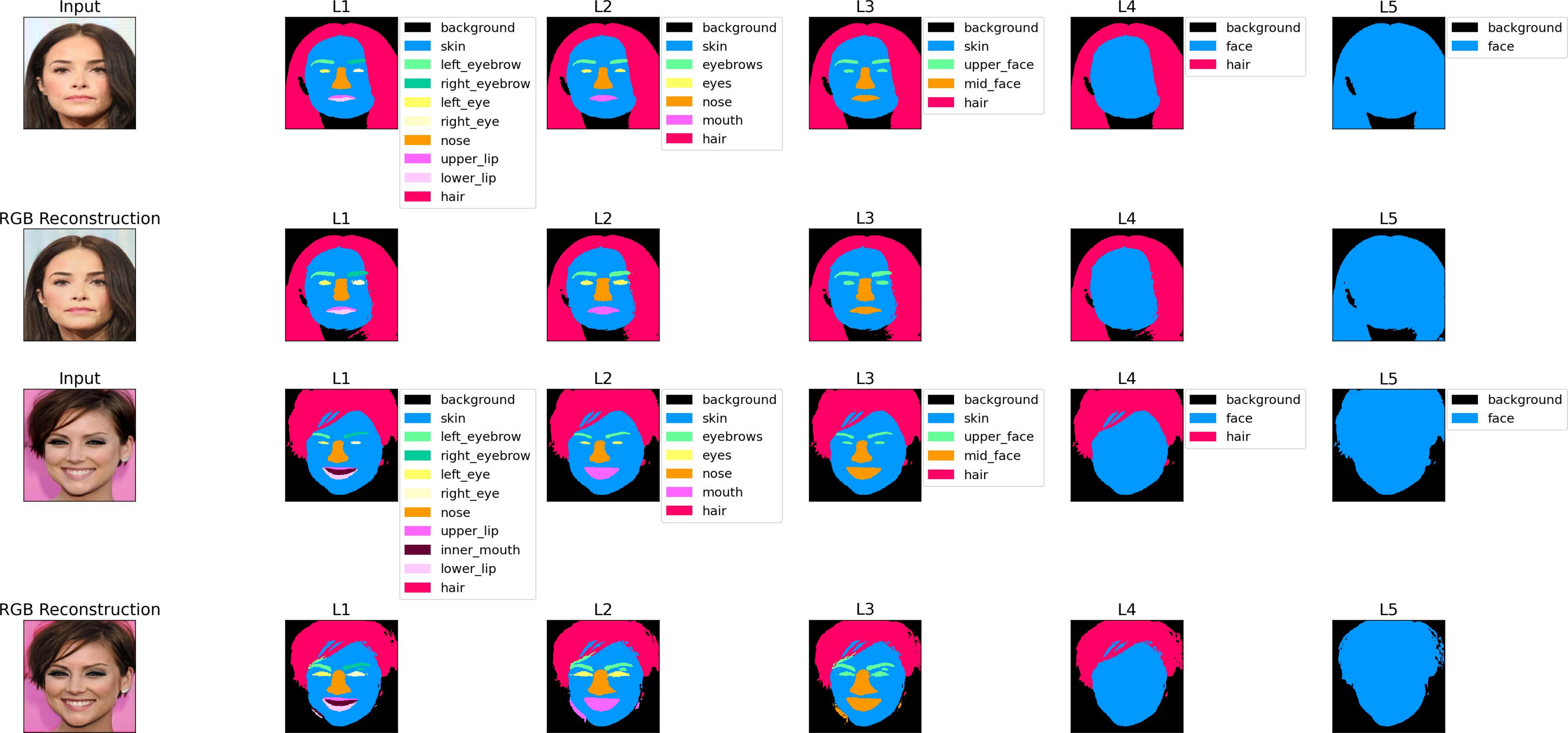}
  \end{tcolorbox}%
}

\twocolumn[
  \icmltitle{
  Modern GLOM \\
  Canonical Locks that encode part-whole hierarchies
  }

  \icmlsetsymbol{equal}{*}

  \begin{icmlauthorlist}
    \icmlauthor{Rajat Modi }{yyy}
    \icmlauthor{Yogesh Singh Rawat}{yyy}
    % \icmlauthor{Geoffrey Everest Hinton}{equal,yyy}
    %\icmlauthor{}{sch}
    %\icmlauthor{}{sch}
  \end{icmlauthorlist}

  \icmlaffiliation{yyy}{Department of Computer Science, University of Central Florida, United States}

  \icmlcorrespondingauthor{Rajat Modi, Yogesh Singh Rawat}{rajatmodi62@gmail.com, yogesh@crcv.ucf.edu}
  \icmlkeywords{Machine Learning, ICML}

  \vskip 0.2in

  \begin{center}
      \usebox{\teaserbox}
      \captionof{figure}{Prediction of part-whole hierarchy of a face from Asynchronous Perception Machines. It contains 5 levels (L1-L5), going from finest to coarse grouping. The net generalizes to any unseen image.}
      \label{fig:celeba_hierarchy_preds}
  \end{center}

  \vskip 0.2in
]

\printAffiliationsAndNotice{}  % no special notice (required even if empty)

\begin{abstract}

 One of the challenges in representational learning is how to encode part-whole hierarchies in a neural net. Prior works rely on flattening tree-like structures into string-like sequences and training a sequence-to-sequence model via autoregression. While such a representation works for parse-trees in NLP, it is not entirely clear how to make it work for images. Thus, we propose a geometric primitive called canonical locks. The key idea is that parts/wholes can be modelled as higher-dimensional vectors ($d \geq 4$), and information can be encoded in their relative phase differences.

  Inductively, the net consists of positionally-bound bottom-up and top-down neural fields, which drive each other to achieve a state of thermal equilibrium. Additionally, we show the existence of a few symmetrical configurations in the net. The computational iterations taken to break these symmetries depend on the angle between parts/wholes arranged on a disk (or more precisely a ring) in higher dimensions. It also appears to have connections to the psychological phenomenon of mental rotation. 
\end{abstract}

\vspace{-2.5em}

\section{The Problem of Part-Whole Hierarchies}

Most of the data we encounter in real world, can be modelled as a graph. For example, a molecule can be modelled as a graph of atoms, social networks can be modelled as graphs of people, and even the brain can be modelled as a graph of neurons. More importantly, such graphs often change dynamically according to the input. For eg, a face can be modelled as a graph of parts (eyes, nose, lips, etc), but the structure of this graph changes if the face is rotated or occluded. Similarly, most of the times, this structure is not known apriori, and must be inferred from the input.

This raises the question: how can a neural net encode part-whole hierarchies whose structure can change dynamically given a particular input? Indeed, the importance of this problem has been further articulated at great length  by Geoff Hinton in his 2021 proposal for the GLOM architecture\cite{glom}. However, an implementation for GLOM has continued to remain elusive for our community.

Over the years several attempts have been made. For eg, \cite{Modi2024OnOI} showed the existence of islands of agreement inside a video-transformer trained only on recognition\cite{fan2021multiscale}, which was a key theoretical representation in Hinton's GLOM. Recently, it was also suggested that global self attention does not allow part-whole hierarchies to emerge in a transformer and instead inverts them. In ICML 2026, \cite{liu2026krause} presented Krause synchronization transformers, a form of constrained attention that  appears to promote the  cluster formation .

 It was later realized that GLOM might require a new fundamental computational unit (different from McCulloch-Pitts neurons) which could promote cluster formation in a neural net. For eg, Takeru (and collaborators) \cite{miyato2025artificial} invented a new kind of neuron called AKORN which relies on principles of neural synchronization. Similarly, Sindy \cite{lowe2023rotating} presented the idea of rotating features as an alternative to slot based representations.

Concurrently, \cite{modi2024asynchronous} demonstrated a potential implementation of GLOM using the asynchronous perception machine (APM), and a potential validation of GLOM's idea that percept is indeed a field. Of note was their invention of a geometrical operator called `the folding-unfolding operator', which is analogous to the phenomenon of mental folding if one models the brain as an analog memory\cite{hinton1979some}. 

Such progress over these past five years motivates us to ask ourselves (yet) again: Could the ideas promised in GLOM/APM be  made to somewhat work? Could what once seemed like a philosophy now lie firmly in our grasps? Indeed, there are three issues with APMs which prevent the philosophy from becoming a  reality. Notably:
% \vspace{0.2em}

\begin{itemize}[itemsep=0.1em]
    \item APMs only works for test-time-training. For purely supervised task, for eg, image classification, the internal representations of the net were severely collapsing. It was possible to average them and perform classification (using an external teacher's vision encoder), but attempts to do dense segmentation were met with utter failure. 
    \item Furthermore, applying regularization to different cortical columns in APM to overcome collapse, applying contrastive learning on local patches of a image, or forcing the net to reconstruct masked regions of an image proved too slow to converge. To give an estimate, it took almost a day to train a cifar-10 classifier from scratch including the pre-training phase\cite{garau2022interpretable}. 

    \item Later on it was realized that codistillation across locations from a strong teacher (eg Dinov2) got the training time down to a couple of hours \cite{modi2024asynchronous}. While APM showed that the net could distill last layers of the teacher, the paper did not consider the subtle matters of part-whole hierarchies. There was also the obvious flaw, that it relied on a teacher, which in turn had to be trained at scale. 

    \item A further assumption APM made was that internal layers of a transformer naturally encoded part-whole hierarchies. However, that quickly proved to be wrong. 
\end{itemize}
% \vspace{0.2em}

To resolve some of the above issues, we propose the second generation of Asynchronous Perception Machines (APM/s). Of note is the ability of APM/s to learn only on one train sample (without any pre-training), and generalize to an entire test set with competitive accuracies. APM's also appear capable of encoding part-whole hierarchies, as well as finally doing dense-segmentation tasks. The decoded features are also robust to occlusions and masking.

This paper is an attempt to interpret GLOM's philosophy into engineering constraints which may in turn be implemented onto parallel distributed memories. A subsequent paper (currently under drafting) shall present our quantitative findings in a clear and succinct manner. Our sole purpose here is to  lay out a language of vector spaces in which one can begin to think about parts, wholes, their compositions, and decompositions. The terminology established here will be used in our subsequent papers. 

We begin by discussing two ways of encoding part-whole hierarchies in neural nets.
\section{Two ways of encoding part-whole hierarchies }

\begin{figure}[ht!]
    \centering
    \includegraphics[width=0.5\textwidth]{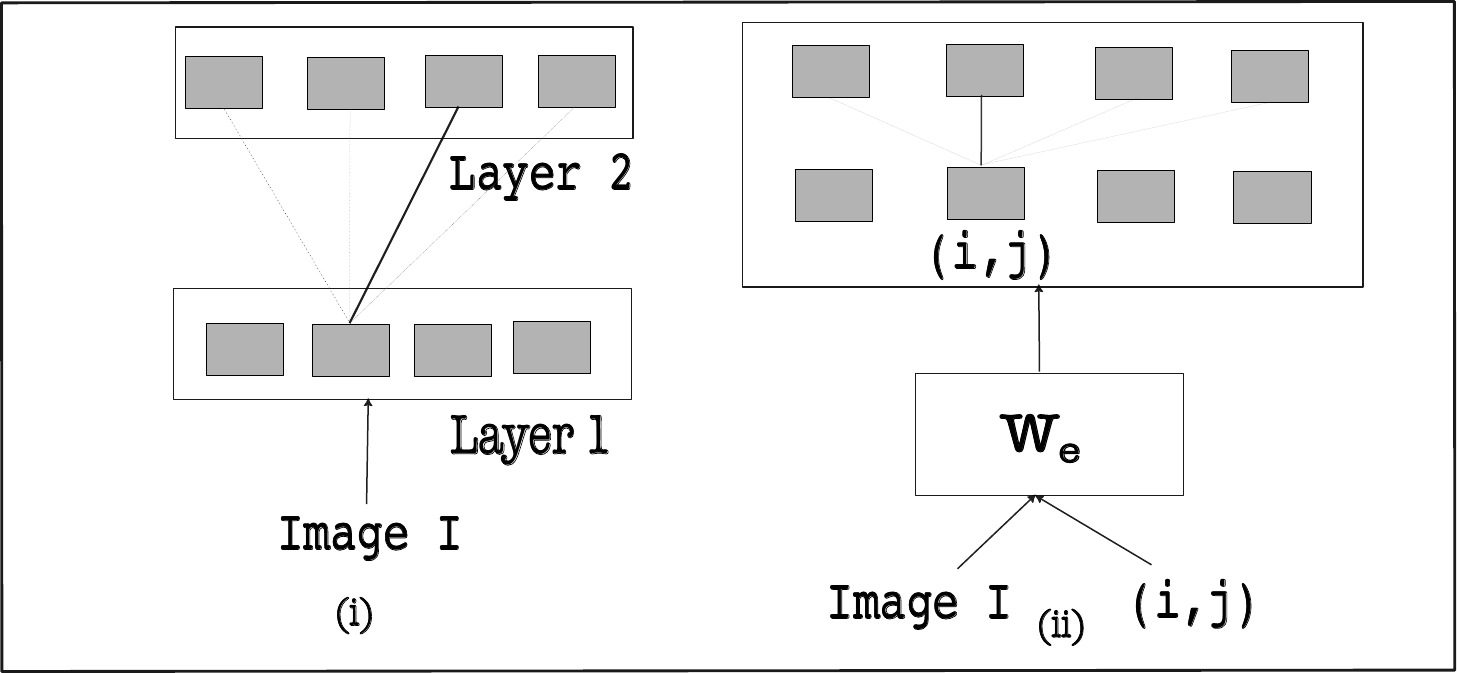}
    % \vspace{-1em}
    \caption{Two ways of encoding part-whole hierarchies.  (i) Encoding hierarchy inside the neural net; (ii) Encoding hierarchy outside the neural net. A darker color on the neural connection indicates a higher probability that a given part chooses a whole in a higher layer as its parent.}
    % \vspace{-1.9em}
    \label{fig:background}
\end{figure}

% For those who choose to stay, we discuss several ways of encoding part-whole hierarchies.
\subsection{Encoding hierarchy inside the net} Fig\ref{fig:background}(i) illustrates a hierarchy with a sequence length of four, and two levels. An image can be fed through several attention layers (as in a ViT), without loss of spatial resolution. Each layer generates internal representations (marked as grey tokens).  One can encode part-whole hierarchies by taking a dot product between parts and wholes, and calculating attention weights (after squeezing logits through a softmax bottleneck). The resultant matrix can be forced to be the same as a graph's adjacency matrix. A similar coding scheme was used in capsules\cite{sabour2017dynamic}. However, this raises several interesting issues:

\vspace{0.2em}
\begin{itemize}[itemsep=0.1em]
    \item Routing issue: Each part $p_i$ must learn which whole $w_j$ it belongs to. For $N$ parts and wholes, this represents an $N^2$ assignment problem. All this routing information becomes noisy as data passes through different layers, which cannot be compensated for by dynamic routing\footnote{which acts as a single feedforward approximation of an EM iterative clustering algorithm as in DeepSets\cite{zaheer2017deep}}. Stacking several such layers appears to make the problem worse.
    
    \item Different graphical structure across each layer: The adjacency matrix of part-whole relationships changes across different layers of the net. Encoding this dynamic evolving structure is more challenging than enforcing the same graphical structure through different attention layers\cite{velivckovic2017graph}. 
    
    \item Collapse issue:  The representations produced in higher layers tend to become identical across different wholes, which means that the net cannot reliably encode multi-level hierarchies. 
\end{itemize}
\vspace{0.2em}

Typically it is possible to sweep these issues under the rug by pre-training a transformer-like net on vast swaths of data. The scale of data drowns out the natural noise of the system, and thus forces it to learn. At the moment, there are two philosophical schools of thought. Although the connectionist-symbolist war has been (somewhat) pacified and connectionists have won, a new kind of war has brewed in the shadows:

\vspace{0.2em}
\begin{itemize}[itemsep=0.1em]
    \item On one side, we have Sutton's bitter lesson: It states that we should stop thinking about contents of minds in terms of simple ways like objects and symmetries. 

    \item On the other side we have Geometrical school of thought: which says that  canonical frames, mental rotations, part-whole hierarchies, and equivariance ought to be a fundamental component of representational learning.   

\end{itemize}
\vspace{0.2em}

Indeed, geometrical school of thought is older, for it has been around since the time of Plato. The scaling school of thought is newer, for computing machines have only recently gotten faster. Things that are new, seem shiny, and move faster. Things that are old are heavier, and move slowly. The momentum thus is product of mass and velocity. 

Now that the warring factions have been defined, we must continue on. But, you dear reader, must keep your mind open, and judgement neutral, for these are the times of peace. Those who render judgement too quickly are forced to choose one side, and thus only win with a 50\% probability. Those who however negotiate peace, and GLOM together everything\footnote{GLOM is derived from the slang ”glom together” which may derive from the word ”agglomerate”\cite{glom}}, have a 100\% percent chance of winning, for no man (or woman or transformer\cite{Vaswani2017AttentionIA}) shall be left behind.  

From a biological standpoint, we are still not able to explain why our minds are able to consume so less energy and exhibit vast abilities of generalization. The explanation that generalization emerges from scale does not explain how we as humans, in our limited lifespans, with our limited experiences, can generalize to vastly different scenarios.Indeed, it seems as though we are missing something fundamental.

Our results suggest that it is indeed possible to combine the benefits of scaling with that of faster search: by constraining the neural net by the laws of geometry. For this unification, we introduce this notion of encoding part-whole hierarchies `outside' a neural net.

\subsection{Encoding hierarchy outside the net}
\label{sec:outside_net}
Fig \ref{fig:background}(ii) shows an alternate coding scheme. First, a neural net (with weights $W_{e}$) takes an image $I$, a location $(i,j)$ as input and produces an embedding. For a sequence of length $L$ and $N$ levels, there are $N*L$ embeddings. The neural net $W_{e}$ could be a  multi-layered perceptron with non-linear activations or a vision transformer (ViT). This kind of model has several advantages over the former:
  
\vspace{0.2em}
\begin{itemize}[itemsep=0.1em]
    \item Embeddings at different locations $(i,j)$ can be computed in parallel, which makes it faster and possible to spread queries across multiple GPUs. 
    \item For a given query $(i,j)$, all the neurons in $W_{e}$ are used to compute the embedding at that location. The binding is now done by a location-specific query.
    \item The amount of available neural hardware does not change and remains reusable  no matter how deep the hierarchy gets. 
\end{itemize}
\vspace{0.2em}

This coding scheme of encoding hierarchies in the latent space outside the net was  proposed by \cite{glom}, and later on inductively realized by \cite{modi2024asynchronous}. However, it collapses when multiple levels of the hierarchy are considered. This leads us to the notion of canonical locks\cite{hinton1981parallel}.

\section{Canonical locks for learning machines}
We begin with a toy problem. Next, we consider two kinds of computational systems: (i) static feed-forward inference; (ii) dynamical systems like a net of oscillators or continuous Ising machines. Inspired by \cite{sejnowski1986learning}, we isolate several symmetrical configurations in the net and demonstrate how they are resolved.  Simultaneously, we examine the connections between canonical locks and mental rotation. 

\subsection{A toy problem}

\begin{figure}[ht!]
    \centering
    \includegraphics[width=0.4\textwidth]{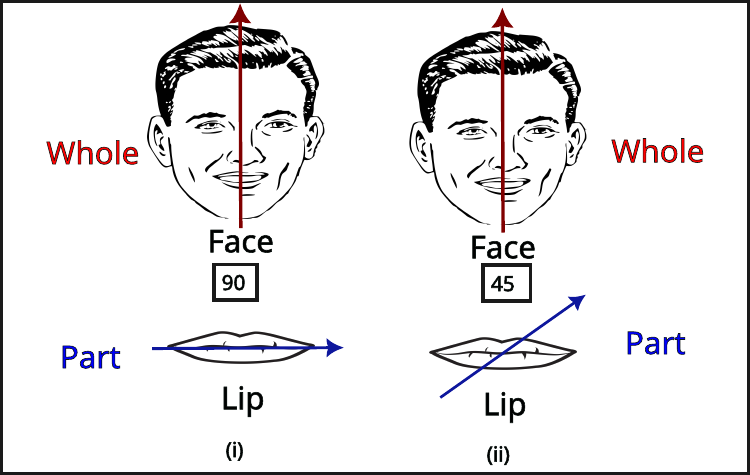} 
    \caption{(i) Face is at $\mathtt{90}^{\circ}$ w.r.t lip; (ii) An alternative hypothesis where face is at $\mathtt{45}^{\circ}$ w.r.t lip. Visually, there is no constraint about which of the angles $\mathtt{45}^{\circ}/\mathtt{90}^{\circ}$ will be preferred.}
    \label{fig:my_svg}
\end{figure}

Fig \ref{fig:my_svg} shows a face (whole) and one of its parts (nose). We can draw little arrows through them. The direction of each arrow governs its orientation. For example, in (i), the lip is at $\mathtt{0}^{\circ}$ and the face at $\mathtt{90}^{\circ}$. Similarly, in (ii), the relative angle between the face and lip is $\mathtt{45}^{\circ}$. This arrow can be a vector in $\mathbb{R}^d$. A neural net won't be able to learn anything useful if the relative angles between the whole and the part keep changing. It needs to be able to select one among infinitely many such angles. First, we present a way to visualize these vectors.

\begin{figure}[ht!]
    \centering
    \includegraphics[width=0.4\textwidth]{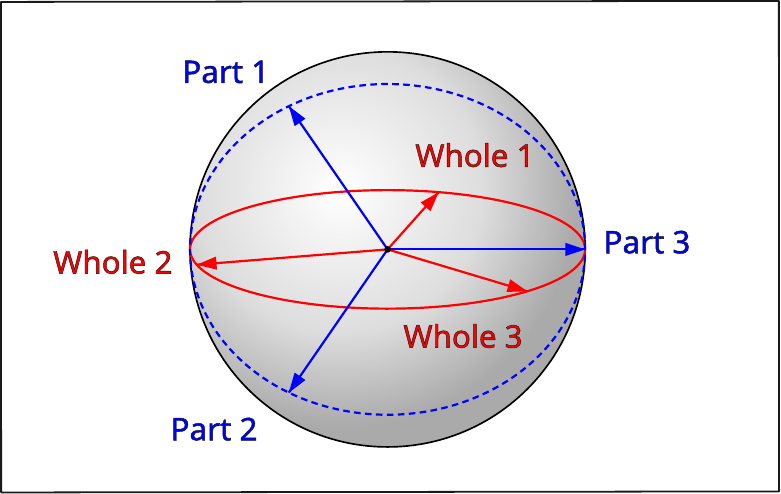} 
    \caption{A hypersphere analogy. Each disk is a great circle of the largest possible radius lying on the sphere.}
    \label{fig:lock}
    \vspace{-1em}
\end{figure}

Fig \ref{fig:lock} shows a hypersphere of unit radius. We imagine two disks on the surface of this hypersphere. One disk corresponds to the part vectors, and another to whole vectors. A vector connecting the center of the sphere with the circumference of the disk represents a part/whole vector. At any moment, we are allowed only `one' vector per disk. 

For example, the whole's vector can oscillate between one of the states like whole 1, whole 2, whole 3, etc. Similarly, the part's vector can oscillate between part 1, part 2, part 3, etc. There are infinite such states, and a vector can be in any of them at a given time. This representation then starts to bear resemblance to continuous Ising models\cite{glom}. 

Please note that a vector possesses a magnitude $r$ and a phase $\theta$. This does not encode in which direction the vector is spinning (clockwise or anti-clockwise). For some tasks, it may be required, but for our toy problem, we will ignore this (subtle matter of chirality of the digital organism). For the rest of this paper, instead of looking at the sphere (which is tough  to draw), we will visualize the disks themselves.

\begin{figure}[ht!]
    \centering
    \includegraphics[width=0.4\textwidth]{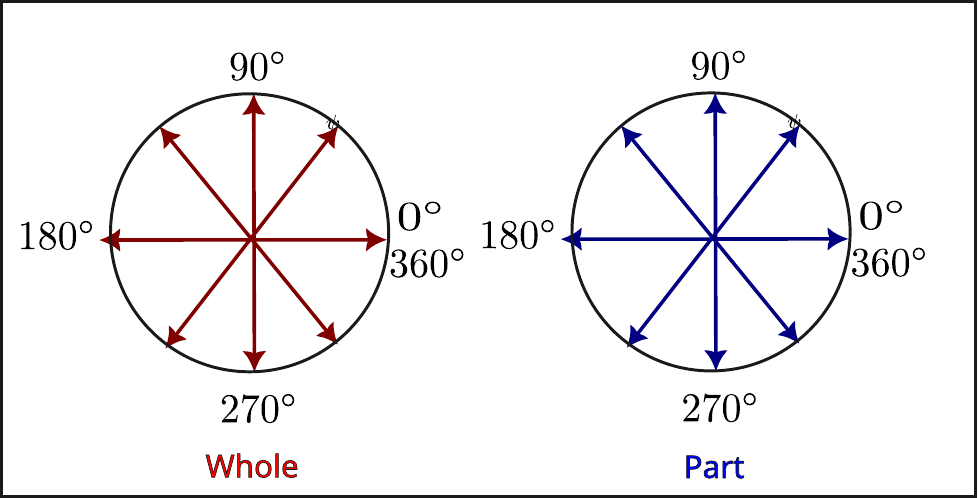} 
    \caption{A 2D projection of the great circle taken from the hypersphere. Part's and Whole's arrows can be thought of as `needles' oscillating in a compass.}
    \vspace{-0.5em}
    \label{fig:disk}
\end{figure}

In Fig \ref{fig:disk}, the visualization now changes. We can think of the part and whole vectors as little needles lying on a circular compass. Any vector is free to choose any angle on the disk. However, if the part and whole are to be related to each other, they need to decide on a particular angle. This is what we now call a canonical lock\cite{hinton1981parallel}.

\begin{figure}[ht!]
    \centering
    % \vspace{-0.5em}
    \includegraphics[width=0.4\textwidth]{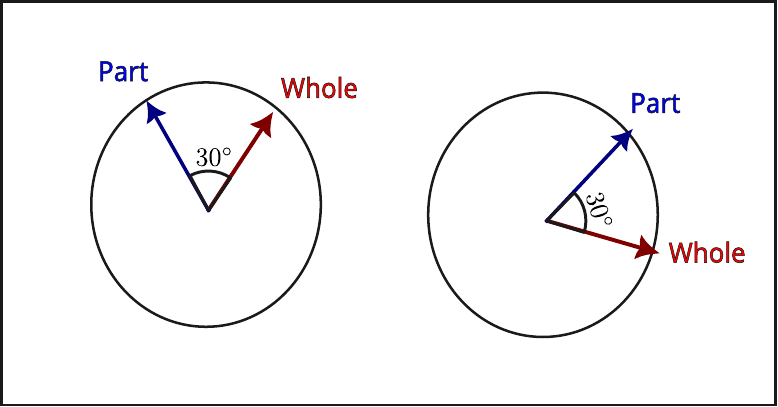} 
    \caption{A perfect canonical lock.}
    % \vspace{-1em}
    \label{fig:canonical_lock}
\vspace{-1em}
  \end{figure}

In Figure \ref{fig:canonical_lock}, the part (lip) is perfectly locked to the whole (face) at an angle of $\mathtt{30}^{\circ}$. The left and right parts of the picture show these vectors `rotating' on the disk in different configurations\cite{lowe2023rotating}. At different instances of time, the part and whole vectors may be at different angles, but the relative angle between them remains constant. The job of a neural net is then to achieve such a `canonical lock'. 

This form of representation raises several concerns:

\vspace{0.2em}
\begin{itemize}[itemsep=0.1em]
    \item For a single pair of $\left\langle \text{part}, \text{whole} \right\rangle$ vectors, what angle should be chosen? This information is not readily available as supervision in the part-whole hierarchy. 
    \item For many such pairs spanning the hierarchy, how can a neural net achieve multiple canonical locks simultaneously? Can simple gradient descent learn this?
    \item If the input vector is rotated (translated) slightly, the predicted representations of the net should also rotate (translate) in an equivariant manner. The canonical locks however should remain invariant.  How can we enforce both equivariance and invariance at the same time\cite{cohen2016group}, in particular, for the attention mechanism\cite{hutchinson2021lietransformer,Vaswani2017AttentionIA}?

\end{itemize}
\vspace{0.2em}

We will try to answer these questions as we proceed in the paper. First, we wish to discuss static vs dynamical systems \cite{lowe2023rotating, akorn}.

\subsection{Static vs Dynamical systems}

We now distinguish between two kinds of recurrent nets in which part-whole hierarchies may be encoded:

\vspace{0.2em}
\begin{itemize}[itemsep=0.1em]

\item Static System: Models like Deep Equilibrium Models\cite{bai2019deep} assume that when a recurrent net is run till infinity, it converges to a fixed representation. In such a system, the part and whole vectors might be fixed in both magnitude and direction once the fixed point is reached.

\item Dynamical System: Here the vectors are given the freedom to rotate on the disk, but they are locked onto each other. Both magnitude and direction of the part and whole vectors may change, but the relative angle between them remains constant. 
\end{itemize}
\vspace{-0.2em}

Intuition suggests that a static system is better: it is easy to detect when the fixed point is reached by comparing the differences among predicted representations across different timesteps of the recurrent net. However, it has been observed that the best downstream accuracy on a given task is not necessarily achieved at the fixed point. Furthermore, running such a net till infinity often leads to loss in performance, a phenomenon dubbed as `overthinking'. 

Alternatively, a dynamical system appears to be more robust. The part and whole vectors keep on rotating forever, but retain their canonical locks. The earlier notion of `stopping' the rotating vectors was an interesting  one: Once the machine reaches a potential solution, it stopped, and the obtained solution was verified by an external observer. If the solution was not correct, it was restarted again.

The uncomfortable decision for allowing the part and whole vectors to rotate for eternity has been made for the following reasons: (i) A system which oscillates can encode timing information as spike trains, which may be useful for tasks involving memory. (ii) Kepler's laws of planetary motion suggest that planets rotate around the sun in predictable orbits; (iii) planets at different distances from the sun sweep equal areas in equal intervals of time; (iv) learning algorithms which allow for conservation of angular momentum then become feasible.

\subsection{Information storage in dynamical systems}

In the forward-forward paper\cite{hinton2022forward}, it was suggested that vectors fed to a subsequent layer of a neural net must be normalized. The reason was that clamping the magnitude to unity results in loss of goodness, which forces the next layer of the net to learn better features. For some reason, separation in the phase space does not affect the downstream accuracy. This assumption makes sense for a static system. 

The system of information organization discussed in Section \ref{sec:outside_net} however is a dynamical system and differs because the last layer of $W_e$ is responsible for generating all the vectors in the hierarchy. It is just a linear layer, and thus free to generate any vector in the latent space, much like CLIP. This means that both the (relative) magnitude and phase of vectors in the hierarchy can be used to encode information. 

A (complex) vector written in the Euler form of $r e^{i \theta}$ means that $r$ is just a scalar, whereas $\theta$ is multi-dimensional. This means that the phase of a vector can encode much more information than its magnitude. The original capsules attempted to encode the presence of an entity in the magnitude. However, this only makes sense if the vector represents a single entity. If an entity possesses multiple identities in superposition, such a representation quickly breaks down. This mysterious case manifests itself when a shape can be either a diamond or a tilted square at the same time. 

Modelling internal representations as complex vectors \cite{lowe2022complex} might allow visualizing different kinds of neural nets one might build, much akin to Smith charts found in electrical engineering. Notably, resistance and reactance can be encoded as a single representation called impedance. The impedance of a circuit of electrical elements is well defined. A neural net is indeed a circuit of parallel interconnected elements, and it should thus be possible to quantify the impedance. However, that matter is a digression, and not relevant to the current discussion. The minds of mortals can only focus on one thing at a time. Multi-head attention does not have those issues, and merely appears to be constrained by memory of the machine.

\section{The Part-Whole Symmetry}
\label{sec:pw_symmetry}

\begin{figure}[ht!]
    \centering
    \includegraphics[width=0.5\textwidth]{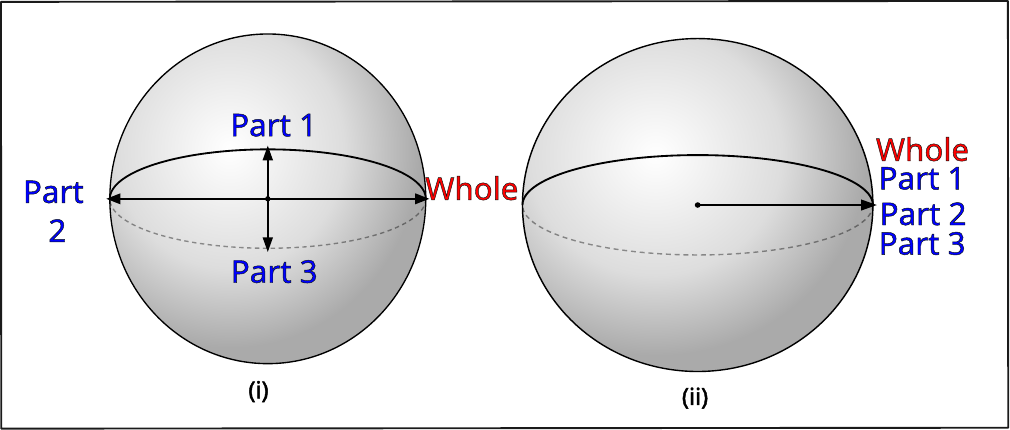} 
    \caption{Some tough symmetrical configurations of parts/whole.}
    \vspace{-1em}
    \label{fig:symmetry_config}
\end{figure}

Now, we consider the case where there is a single whole and multiple parts. For example, in Fig \ref{fig:symmetry_config}, there is one whole and three parts (part 1, part 2, part 3). For the purposes of symmetry, we will illustrate some of their toughest configurations. If we can achieve locking in these configurations, we may be somewhat reassured that the lock may be achieved in other configurations as well.

Figure (i) shows parts/whole arranged perpendicular to each other. If the whole's angle is assumed to be $\mathtt{0}^{\circ}$, then part 1 is at $\mathtt{90}^{\circ}$, part 2 at $\mathtt{180}^{\circ}$, and part 3 at $\mathtt{270}^{\circ}$. Let us assume that the whole vector is kept fixed and the part vectors are refined over several learning iterations. We also assume that the part vectors belong to the same object (for example, the lip). The question is: at what relative angle do we expect the part vectors to converge?\footnote{These assumptions are not necessary, but they help reveal internal mechanisms. As we proceed further, we will try to relax them\cite{hinton1977relaxation}.}

Intuition suggests that the net will choose between part 1, part 2, or part 3 randomly. Alternatively, if we average part 1, part 2, and part 3, the center of gravity lies on part 2. Therefore, we expect the canonical lock to be achieved at part 2. A deeper alternative is that two parts (part 1 and part 3) are at $\mathtt{90}^{\circ}$ with respect to the whole, whereas part 2 is at $\mathtt{180}^{\circ}$. Thus, there is a $\frac{2}{3}$ probability to lock onto part 1 or part 3, and a $\frac{1}{3}$ probability to lock onto part 2.

If we consider the entire system (including part/whole), the center of gravity lies at the origin. One may suspect that the part vector will become the zero vector itself. However, that means that the part vector will have zero magnitude and could possess any direction, rendering any notion of canonical locks useless\footnote{which means we  want it to stay on the hypersphere.}.

An alternate configuration is shown in (ii). Here all parts coincide with the whole. Intuition suggests that the canonical lock is at $\mathtt{0}^{\circ}$. However, this might pose an issue: if all the parts and their wholes are identical vectors, there is no relative angle between them. Thus, the net does not encode any meaningful hierarchical structure.

Till now we have talked about vectors lying on spheres/disks. However, they are typically high-dimensional vectors, sometimes extending up to 2048 dimensions. Recent conversations \cite{akorn} seem to suggest that it is indeed very difficult to achieve a perfect lock in such high dimensions ($d\geq\mathtt{4}$), since the volume of the space is just too high.

While nature does (occasionally) respect symmetry, breaking those symmetries leads to new abilities in the machine. Biological organisms, for example, reproduce and break entropy's law by self-organizing into complex structures. When a zygote reproduces and spreads out in a sphere, a mechanism to break the ring symmetry is what makes our legs look different from our heads. Similarly, we may want our learning machines to break the symmetries and achieve these canonical locks.

\section{The Part-Whole Asymmetry}
\label{sec:pw_asymmetry}

\subsection{Breaking the anti-symmetry}

\begin{figure}[ht!]
    % \vspace{-2em}
    \centering
    \includegraphics[width=0.4\textwidth]{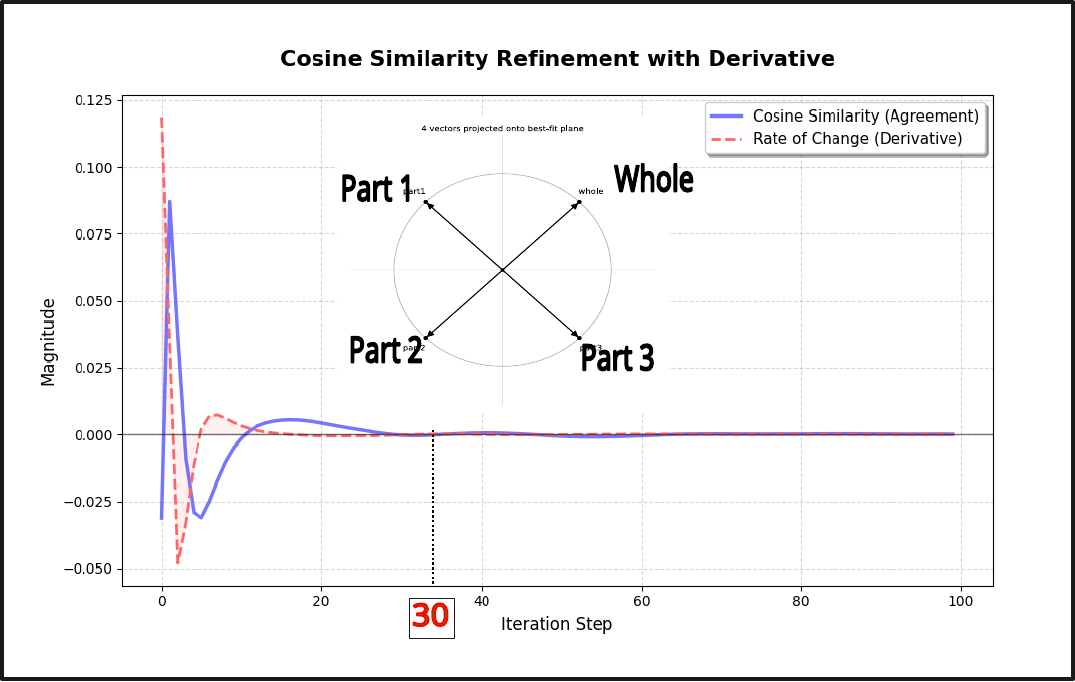} 
    \caption{Resolving the anti-symmetry  }
    \label{fig:circular_symmetry}
    % \vspace{-2em}
\end{figure}

In Fig \ref{fig:circular_symmetry}, we generate 4 vectors which are mutually orthogonal to each other. There is 1 whole, and 3 part vectors. Part 1 and Part 3 are perpendicular to the whole, but part 2 is at 180 degrees. This center of gravity is at 180 degrees with respect to the whole. The assumption here is that the whole vector is held constant, while the part vectors are refined over several learning iterations. The input part vectors are passed through a connectionist net, and the output part vectors are averaged to yield the final part output. We measure the cosine similarity between this part output and the whole vector.

We plot the mean cosine similarity of each part output with respect to the whole. The x-axis is the number of iterations, the y-axis is the mean cosine similarity. We can see that the mutual cosine loss falls to 0 in around 30 iterations. Put simply, given a whole vector, the net discovers an internal representation where all the part outputs are orthogonal to the whole. Note that the net broke this symmetry in $\mathtt{30}$ iterations.

An argument may be raised on the grounds that if one samples any two vectors randomly in higher dimensions, they are likely to be orthogonal to each other. What we then claim as a canonical lock is just a random coincidence. 

The counterargument is that one can build a machine which can be programmed to satisfy any explicit geometric constraints, be they orthogonal or not. If a bunch of vectors are to be arranged in any pattern in the higher dimensional space, the machine shall learn to arrange them.

\vspace{-1em}
\subsection{Breaking the collapsed symmetry}

\begin{figure}[h]
    \centering
    \includegraphics[width=0.5\textwidth]{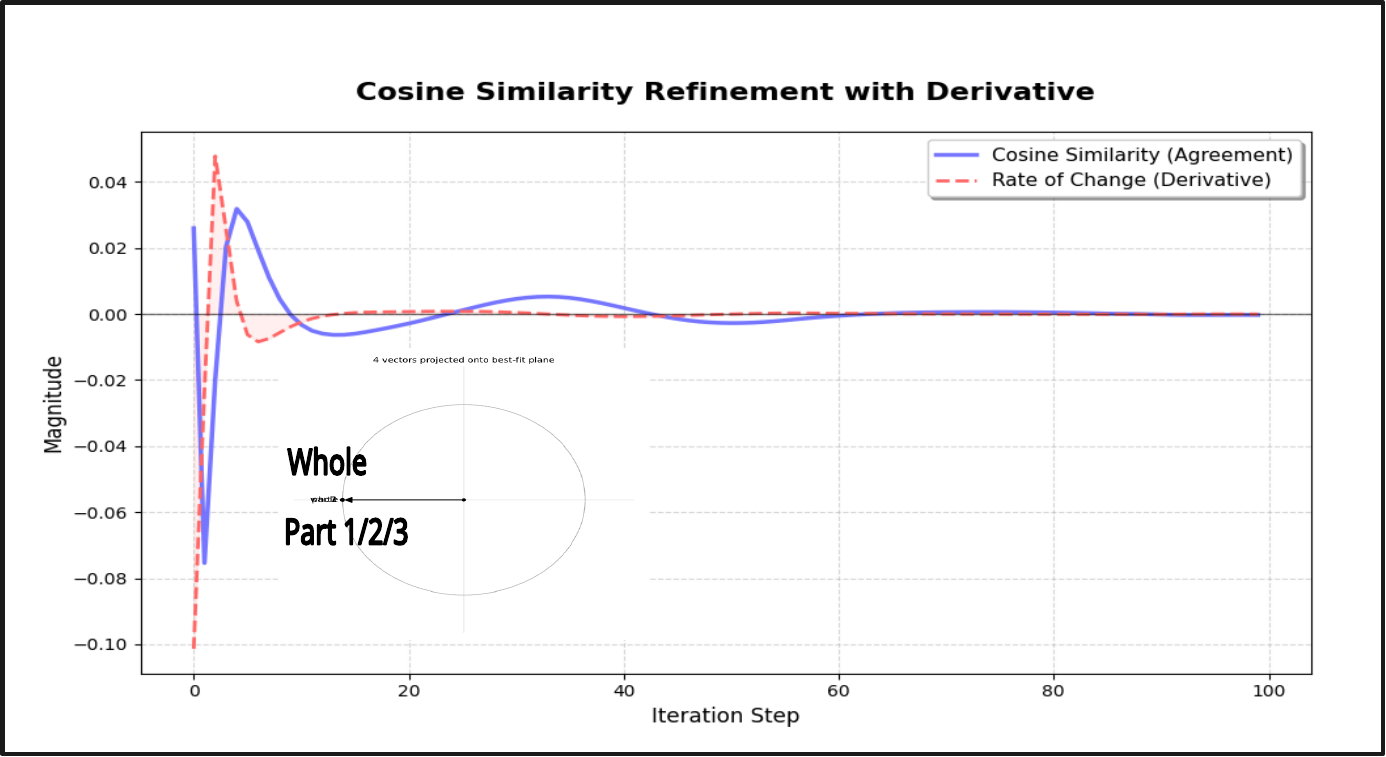} 
    \caption{Resolving the collapsed symmetry. }
    \vspace{-1em}
    \label{fig:collapsed_symmetry}
\end{figure}

In Fig \ref{fig:collapsed_symmetry}, we generate 4 vectors which are identical in both magnitude and direction. There is 1 whole, and 3 part vectors. As before, the net settles on identical part vector outputs which are orthogonal to the whole. This means, even in a state of collapse (when part and whole vectors are identical), the net is able to break symmetry and achieve a lock.

A crucial difference between both cases is the time it takes to break symmetry. In Fig\ref{fig:circular_symmetry}, it took 30 iterations, whereas in Fig \ref{fig:collapsed_symmetry}, it took around 60 iterations. Intuitively, the second case was more difficult since there was no relative structure in part vectors to begin with. This suggests that nature may have built in some configurations that the system may prefer.

\subsection{A Linear Model for Predicting the Time It Takes to Break Symmetry}

An argument made in psychology is that the time it takes for humans to recognize a shape depends on the angle at which they are viewing the shape and the angle of the natural canonical axis they tend to favor\cite{palmer1981cannonical}. Can a neural net follow similar behavior? A demonstration may indeed be performed.

\begin{figure}[h]
  % \vspace{-8em}
    \centering
    % Scaling to 0.48 of textwidth keeps it perfectly inside one column
    \includegraphics[width=0.3\textwidth]{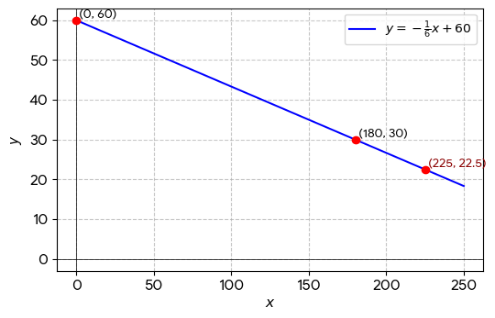} 
    \vspace{-1em}
    \caption{Fitting a linear line model to fundamental symmetries. x axis is no of iterations. y axis is the angle of center of gravity of part vectors w.r.t whole vector.} 
    \vspace{-1em}
    \label{fig:linear_model}
\end{figure}

Consider as before, the case of anti-symmetry in Fig\ref{fig:circular_symmetry} where parts 1, 2, 3 have a center of gravity at 180 degrees with respect to the whole, breaking symmetry at 30 iterations. Similarly, consider the case of collapsed symmetry in Fig \ref{fig:collapsed_symmetry}, where the center of gravity is at 0 degrees with respect to the whole, breaking symmetry at 60 iterations. Thus, we get two pairs of coordinates (0,60), (180,30). We can plot them on a line.

% \vspace{-2em}

% \vspace{-1.5em}

\begin{figure}[ht!]
    \centering
    % Scaling to 0.48 of textwidth keeps it perfectly inside one 
    % column
    \vspace{-1.em}
    \includegraphics[width=0.48\textwidth]{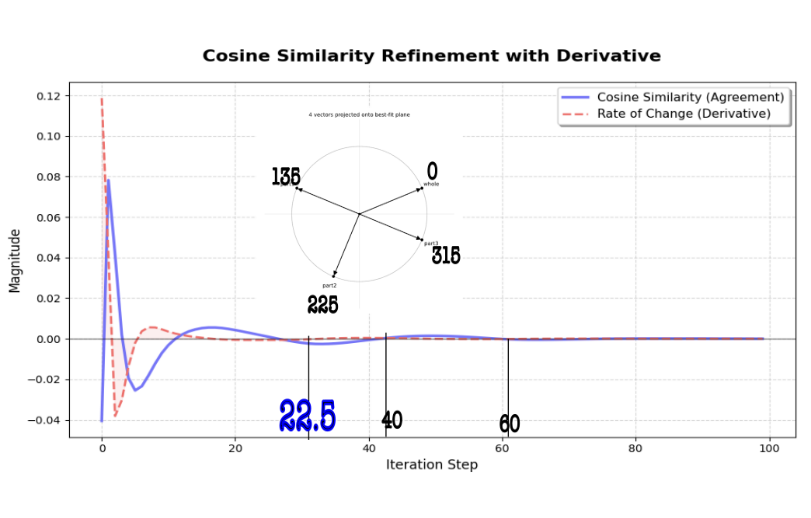} 
    \vspace{-3em}
    \caption{Any random configuration of part-whole vectors on the disk appears to cut $y=0$ multiple times. Each x-axis coordinate (eg,  40, 60) may be thought of as a fundamental symmetry.  }
%   \vspace{-1em}
    \label{fig:225_actual_plot}
\end{figure}

Let us now generate another configuration where parts and wholes are arranged such that the center of gravity is at 225 degrees with respect to the whole. Looking at the linear plot shown in \ref{fig:linear_model}, we predict it should take about 22.5 iterations to break symmetry (rounded as needed). Running a connectionist net with such a configuration results in the graph shown on the next page (Fig\ref{fig:225_actual_plot}).

In Fig \ref{fig:225_actual_plot}, we can see the blue line cut $0$ y-intercept at around 22.5, which is the condition when cosine loss tended to zero, indicating perfect lock. Subsequently, we see the oscillation getting damped, cutting the axis again at iteration 40 and 60. Recall that iteration $\mathtt{60}$ was the case of collapsed symmetry. One unexplained mode is at iteration 40. Looking back at Fig \ref{fig:linear_model}, we see that 40 iterations correspond to a center of gravity of 120 degrees. One way to generate this configuration is as below:

\begin{figure}[ht!]
    \centering
    \includegraphics[width=0.2\textwidth]{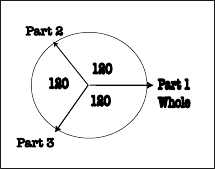} 
    % \vspace{-1em}
    \caption{The tri-fold symmetry }
    % \vspace{-4em}
    \label{fig:225_degree_png}
\end{figure}

Figure \ref{fig:225_degree_png} above represents what we internally have code-named as tri-fold symmetry because all parts are at 120 degrees with respect to each other. The perpendicular components of part 2 and part 3 along the y-axis cancel out, and the center of gravity is at 180 degrees with respect to the whole. However, the true degree of freedom tends to be a bit different. 

We can imagine a plane perpendicular to this paper, and rotate this figure by 120 degrees three times to get the same shape each time. We call this figure of $3$ as a degree of freedom. Dividing the total plausible angle of rotation (360 degrees) by the degree of freedom (3), we get 120 degrees as the smallest angle of rotation that preserves the shape: It is interesting to note that the machine prefers the axis along which this degree of freedom is maximum out of all other axes. We must precisely note what this means. If we want to represent a canonical frame by means of numbers, then all canonical frames look identical. There is nothing in the representation which tells us the degree of freedom the machine possesses given the choice of a canonical frame. 

A way by which the machine can rotate itself along several axis, estimate the degree of freedom along each, and then select the axis along which the degree of freedom is maximum can thus be invented. And it is indeed more elegant than brute-forcing along all the plausible axis. The answer is simply attaching a learnable token to the representation. As long as we break symmetry, the machine shall know what to choose. 

The frame with the highest degree of freedom can yield the same object with the least amount of rotation. Therefore, it is easier for it to learn and only requires minor adjustments to the synaptic weights. However, the number of possible objects for which such higher degrees of freedom exist is very low. For example, the shape of a sphere. This means that the sphere is a Gaussian which has folded all information about the organism and only contains the blueprint for its replication.

This suggests that when the machine is modelling minor rotations of a sphere, it has to be sensitive enough to change its weights to great precision. However, simulating the weights on a digital machine (of 128 bits as of 2026AD) does not offer that precision. It therefore gets really tempting to move onto  qubit/analog kind of continuous Ising machines.

\section{Existence of Other Symmetries}

The three symmetries (anti-symmetry, collapsed symmetry, and tri-fold symmetry) presented above can thus be thought of as fundamental symmetries of learning machines.

Any other configuration appears to oscillate between these fundamental symmetries before finally settling on the solution (a.k.a. equilibrium). The key argument here is that these symmetries emerge even when the net is given random input, initialized with many random weights, and run across multiple seeds.

The assumption till now is that vectors of parts/wholes lie on a disk. There may exist some other configurations where these parts/wholes lie on separate manifolds, giving rise to many more higher-dimensional symmetries. It is very tempting to think of them as fundamental particles the physicists (and their supercool colliders) like to discover every now and then.

\subsection{Synchrony vs Asynchrony across levels of the part-whole hierarchy}

Next, we examine the proposal on GLOM\cite{glom}. Specifically, we are concerned with how information might be represented in the hierarchy. We consider two cases: (i) Synchrony across levels; (ii) Asynchrony across levels. 

\begin{figure}[ht!]
    \centering
    % \vspace{-0.5em}
    \includegraphics[width=0.3\textwidth]{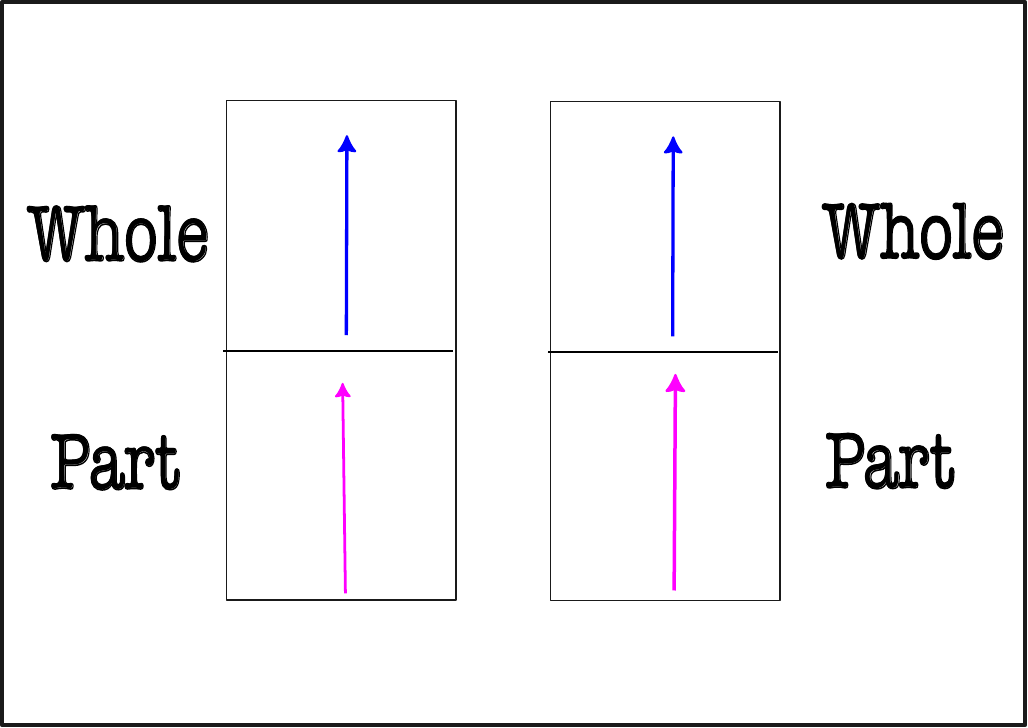} 
    \caption{Synchrony across levels results in loss of information.}
    % \vspace{-1em}
    \label{fig:synchrony}
\vspace{-1em}
  \end{figure}

Fig\ref{fig:synchrony} shows two levels of hierarchy. The lower level has two parts (shown as pink arrows), and the higher level has two wholes (shown as blue arrows). The problem with this case is that vectors across different levels of the hierarchy are synchronized, i.e. they are all at zero relative angle w.r.t each other, and thus collapsed. This means that there is no useful information encoded in this configuration. 

\begin{figure}[ht!]
    \centering
    % \vspace{-0.5em}
    \includegraphics[width=0.3\textwidth]{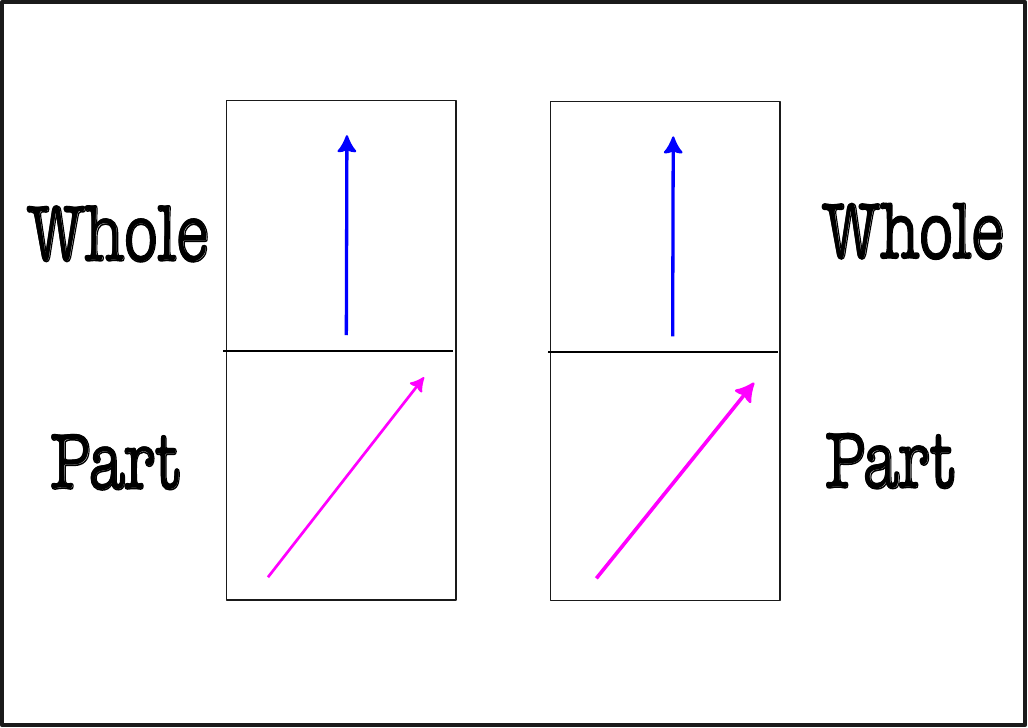} 
    \caption{A successful case of encoding information in a hierarchy.}
    % \vspace{-1em}
    \label{fig:asynchrony}
\vspace{-1em}
  \end{figure}

  Alternatively, consider Fig\ref{fig:asynchrony}. Here, all the parts in the lower level point in the same direction. Similarly, all the wholes in the higher level point in the same direction. However, the relative angle between the parts and wholes is non-zero. This means that useful information has been encoded into the hierarchy. The key takeaway is that relative angles between vectors of different levels of the hierarchy must be non-zero.

\subsection{Relaxing the orthogonality of coordinate frames}

We now consider this notion of coordinate frames as in \cite{hinton1976using}.

\begin{figure}[ht!]
    \centering
    % Scaling to 0.48 of textwidth keeps it perfectly inside one 
    % column
    % \vspace{-1em}
    \includegraphics[width=0.4\textwidth]{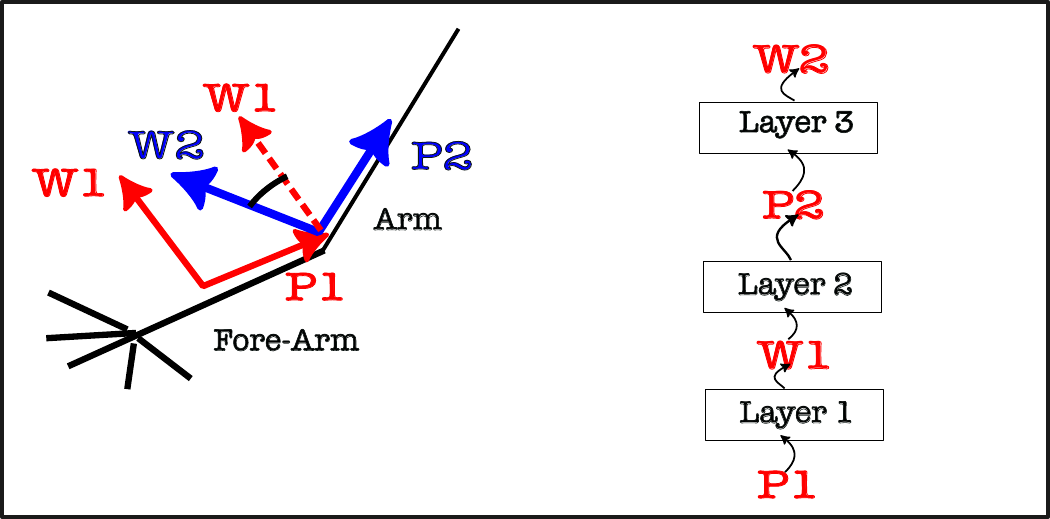} 
    % \vspace{-1em}
    \caption{An arm analogy. Representing coordinate frames with three layers\cite{hinton1977relaxation} }
    \vspace{-1.0em}
    \label{fig:arm}
\end{figure}

Please consider the human stick diagram. It shows two body parts, a fore-arm and an arm. Several observations may be made:

\vspace{-0.4em}
\begin{itemize}[itemsep=0.1em]
    \item There are two coordinate frames, one for the fore-arm and another for the arm. 
    \item Each coordinate frame has an orthogonal basis. For example, the fore-arm has a basis $W_1$ and $P_{1}$, and the arm has a basis $W_2$ and $P_{2}$.
    \item The relative angles between two coordinate frames are encoded as relative angles between the normal vectors of the coordinate frames. For example, the angle between $W_1$ and $W_2$ is $\theta$.

\end{itemize}
\vspace{-0.4em}

It is possible to model this hierarchy by using a three-layered neural net. One can encode a transformation from $P_1$ to $W_1$, then from $W_1$ to $P_2$, and finally from $P_2$ to $W_2$. Note that the only non-orthogonal canonical lock is between $W_1$ and $P_2$. The other canonical locks are all at $90^{\circ}$, which means that they should be easier to achieve. The non orthogonal lock will take some beating to achieve.

\begin{figure}[ht!]
    \centering
    % Scaling to 0.48 of textwidth keeps it perfectly inside one 
    % column
    \vspace{-0.2em}
    \includegraphics[width=0.4\textwidth]{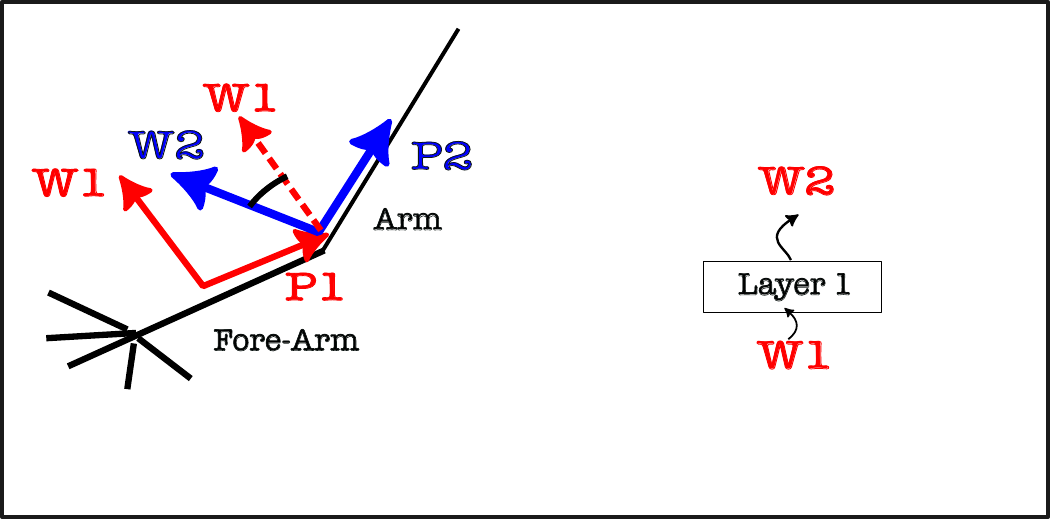} 
    % \vspace{-1em}
    \caption{Representing coordinate frames with a single layer. }
    \vspace{-1em}
    \label{fig:alternate_arm}
\end{figure}

\begin{figure*}[t]
    \centering
    \setlength{\fboxsep}{12pt}   % Inner padding around the image
    \setlength{\fboxrule}{0.8pt}  % Thickness of the border line
    \fbox{%
        \includegraphics[width=0.3\textwidth]{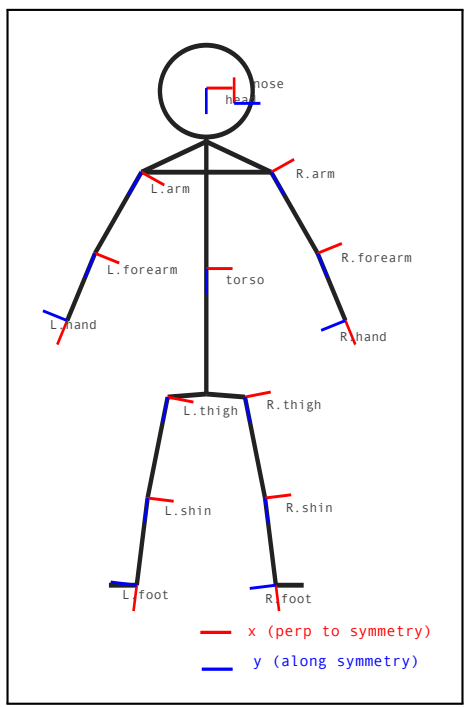}%
    }
    \vspace{10pt} 
    \caption{A human stick figure showing the hierarchy of body parts. Each part has chosen its own canonical axis. Similarly, the part corresponding to the left hand knows that its whole is the left forearm. We want a neural net to learn similar hierarchy.  Figure gratefully inspired from \cite{hinton1977relaxation}.}
    \label{fig:stick_symmetry}
\end{figure*}
There are issues with this kind of representation. The neural net is wasteful, because to model two coordinate frames, we require three layers. In principle, it is possible to model the hierarchy with just one layer.

The idea is to just model the relative angle between the normal vectors of the two coordinate frames, and discard the vectors along the body parts ($P_1, P_2$) entirely. This reduction in the amount of neural hardware may be termed as `relaxing the orthogonality of coordinate frames'. 

A canonical lock between a between a single part and a single whole can then result in canonical locks across the entire hierarchy. Fig\ref{fig:stick_symmetry} shows a human stick figure, where each part has chosen its own canonical axis. Similarly, the part corresponding to the left hand knows that its whole is the left forearm. We want a neural net to learn similar hierarchy within itself.

\subsection{Preserving the orthogonality of relaxed coordinate frames}

Let us now imagine that the angle between $W_1$ and $W_2$ is $60^{\circ}$. This raises a curious problem: forcing a neural net to achieve this canonical lock will consume a lot of energy. 
 For simplicity, we will call the vectors $W_1$/$W_2$ as $v_1$ and $v_2$. Next, we shall  borrow a theorem from higher-dimensional linear algebra.

\begin{theorembox}
\begin{theorem}[Near-Orthogonality of Random Vectors on a hypersphere]
\label{thm:orthogonality}
Let $v_1, v_2$ be independent random vectors drawn uniformly from a unit hypersphere $S^{d-1} \subset \mathbb{R}^d$. Then for every $\epsilon \in (0,1)$,
\[
\mathbb{P}\left(|\langle v_1, v_2 \rangle| \geq \epsilon\right) \;\leq\; 2\exp\!\left(-\frac{d\epsilon^2}{4}\right) + \exp\!\left(-\frac{d}{16}\right).
\]
In particular, for any fixed $\epsilon > 0$, this probability decays exponentially fast in $d$; equivalently, $\langle v_1, v_2\rangle = O_P(1/\sqrt{d})$, so two independently and uniformly sampled directions in $\mathbb{R}^d$ become orthogonal with overwhelming probability as $d \to \infty$ \hfill$\blacksquare$

\end{theorem}
\end{theorembox}

The above theorem, although obvious in hindsight , does carry several interesting implications for connectionism.

\vspace{-0.2em}
\begin{itemize}[itemsep=0.1em]

\item As the dimensionality of a vector increases, it can encode a lot more information. In principle, a single vector could carry information about an entire scene, and it is possible to decode the entire part-whole hierarchy from it. 

\item Most of the vectors in a higher-dimensional space are orthogonal to each other. 
\end{itemize}
\vspace{-0.2em}

A neural net in which the vectors are almost orthogonal to each other will need many computational iterations to drive the internal canonical locks to $60^{\circ}$. It may be possible to adjust the weights of the net in one rapid update, but that would require making changes to the perceptron convergence procedure.

An alternate way is not to fight the fundamental laws of physics and try to force structure in the net through iterations of gradient descent. Rather, one must embrace the fact that the vectors `will be' orthogonal. Submitting to the laws of nature must happen, if we are to crack the energy barrier in which the current generation of intelligent machines is trapped. Biological beings however repeatedly beat the laws of entropy. Soon after their passing, however, the flow of energy gets reversed.

This leads one to the question: Is there a way to arrange these vectors together into some sort of geometric structure, so that a combination of them could encode part-whole hierarchies?

The motivation stems from trying to resolve Fodor's arguments against connectionism\cite{fodor1988connectionism}. It has been shown that symbols (characters) and their combinations (words) can be encoded as vectors in neural nets. Similarly, image pixels can be encoded as vectors. However, it is not clear how pixels form objects, for there is no definition of what an object is. In the words of David Marr: `what is an object?'
It merely depends on the granularity at which one is observing them. 

This further suggests that neural hardware must be designed to abstract higher-order mental functions at different levels of granularity, for eg, representations at level one could also serve as level ten at some other scale. After all, we have no trouble in looking up at the galaxy of stars, or the world of atoms in a microscope. The universe is countless at different scales, which might suggest that there is no informational bottleneck. 

If one is willing to accept the premise that neural nets operate on higher vector spaces, two arguments resurface:

\vspace{-0.2em}
\begin{itemize}[itemsep=0.1em]

\item How do vectors arrange themselves to form concepts like `face', `lip', `nose' etc. in a higher-dimensional space? Is there a geometric structure which can be imposed on these vectors, so that they can encode such part-whole hierarchies? If there are infinite concepts, how can they be coded onto finite neural hardware?
\item How are concepts represented as distributed patterns? Is it possible to learn them as geometric primitives, whose combination can lead us to the formation of new concepts on the fly? Is it possible to capture these primitives as continuous probability distributions one might sample from? The structure of these hierarchies should be changed dynamically based on the input data vector, and not be fixed a priori.
\end{itemize}
\vspace{-0.2em}

Indeed, these arguments allow us to probe whether the brain relies on a universal form of grammar of symbols\cite{chomsky1965aspects}. Does an organism get born with such a universal grammar (or somehow inherit it through the genetic code from its parents), or does it learn it from the environment as it grows up\cite{hinton1987learning}? If the latter were true, then an organism isolated from the environment should not be able to learn any language. 

An alternate possibility is that the language transmitted by the parent is a universal language of syntax, and entities which satisfy it lead to the formation of higher-level concepts, which allows one to encode causal constraints. A rule that A:B gives A+B should hold no matter what A and B are.

This suggests that if a neural net wishes to express the rule of A:B$\rightarrow$ A+B in its weights, it needs to adjust the configurations of its internal part-whole vector hierarchy accordingly. A and B are then input data vectors presented to the bottom of the net. The part-whole hierarchy operates on these vectors for certain iterations, undergoes a settling/relaxation process, and then produces the final answer. This then brings us to the notion of a part-whole grammar.

\subsection{The part-whole grammar}

\begin{figure}[ht!]
    \centering
    % Scaling to 0.48 of textwidth keeps it perfectly inside one 
    % column
    \vspace{-0.2em}
    \includegraphics[width=0.4\textwidth]{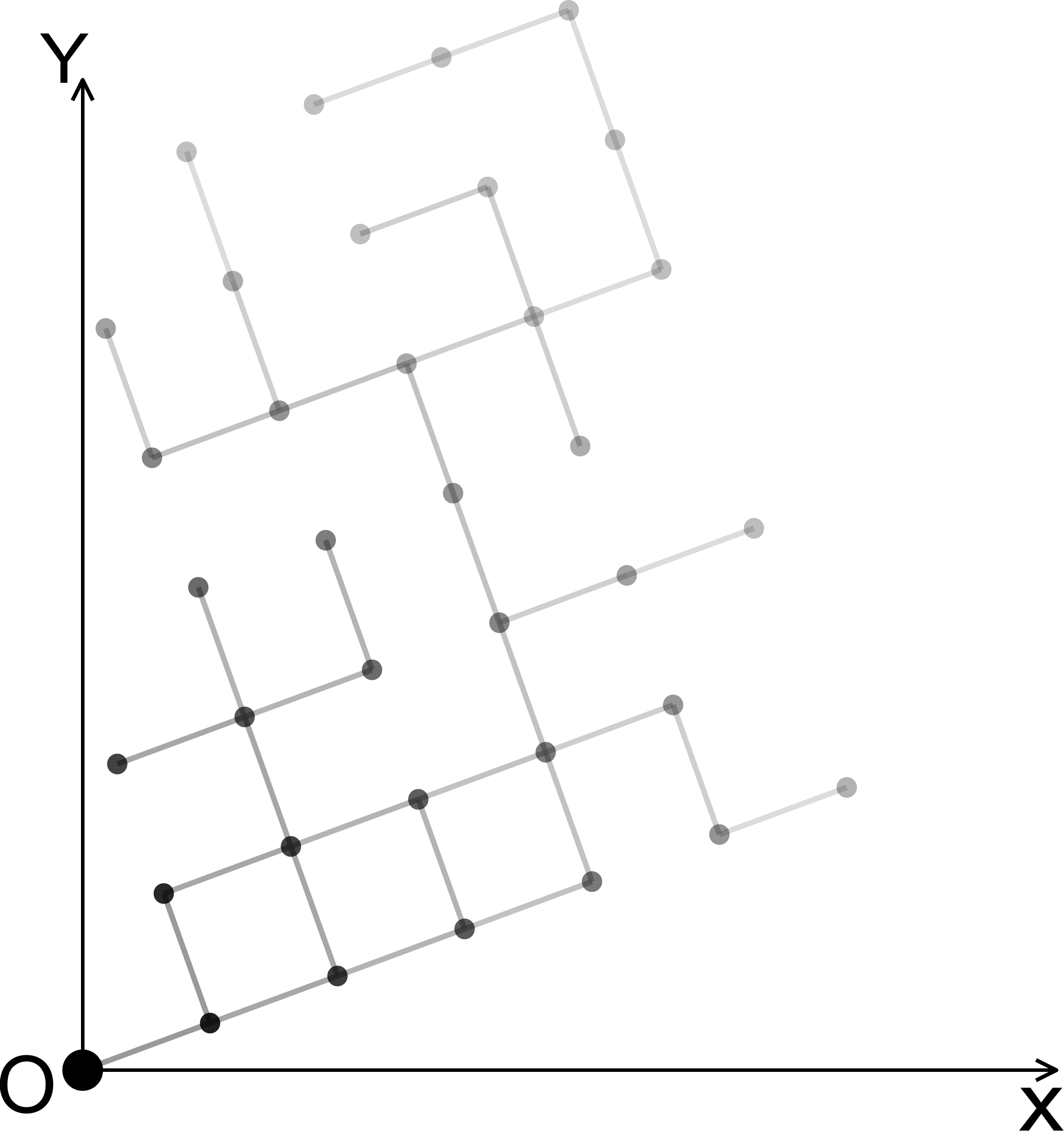} 
    % \vspace{-1em}
    \caption{Lattice of a learning machine. Solid lines are connections already formed. Dotted lines are connections currently being formed. The lattice is equivariant around the origin. }
    \vspace{-1em}
    \label{fig:lattice}
\end{figure}

 Fig \ref{fig:lattice} reveals a model of how representations may arrange themselves as a hierarchical structure inside a higher-dimensional space. The machine operates in a few phases. 

 First, the internal representations of the net are purely random and have no structure. Given a problem of interest, the net starts to grow a lattice-like structure in a higher-dimensional space. The primitive building blocks of this lattice are vectors and disks. The actual learning mechanism is how to arrange these vectors and disks in a particular configuration, so that they can represent an appropriate part-whole hierarchy. 

 The picture shows us one such lattice being formed. The solid lines represent connections which have already been formed, whereas the dotted lines represent connections which are currently being formed. Kindly note that the lattice is equivariant around the origin, which means that if the entire lattice is rotated around the origin, it will still represent the same part-whole hierarchy.

 The knowledge of a neural net is the set of weights that produce a particular lattice configuration. Intuitively, for a given problem of interest (or an algorithm), there is a unique lattice that encodes the `solution' to that problem. For multiple problems, the net will need to learn multiple lattices. This means that the neural computation may proceed  in the following manner:

\vspace{-0.2em}
\begin{itemize}[itemsep=0.1em]

\item First, a symbolic algorithm (like sorting an array) is presented to the net as a way to solve a problem.
\item The net encodes it as a mere lattice configuration in the higher-dimensional space. An indicator that a lattice has been formed is that the part-whole vectors achieve perfect canonical locks, which can be measured by looking at the internal energy of the net between two successive iterations. Forming the lattice should require no example of the problem being solved. 
\item Once the lattice is formed, the net can then be presented with a problem instance (like an array of numbers to be sorted). The net will then use the lattice configuration to solve the problem. The solution will be produced as a vector which can be eventually decoded into a symbolic answer.  
\item The structure of the lattice does not change with different instances of the problem
\end{itemize}
\vspace{-0.2em}

This notion of a field of lattice reduces neural nets to mere hardware that implements higher-order symbolic logic. This implies that connectionism has to bow down to the rigor of formal logic, for it still cannot solve problems of binding and compositionality properly, and remains a mere statistical correlation machine. However, there is no need to feel ashamed of that, because there may be a way to achieve peace between the two schools of thought. To understand the differences, we
 must first properly partition the problems we might wish our dear computing machines to solve for us. 

\uline{Problems whose algorithms exist:} Let us imagine the problem of sorting numbers. A sequence of steps to achieve the task is already known. If one is given the algorithm, he does not need any instance of the problem to `learn it'. What one desires is the ability to encode the algorithm in weights of the distributed memory of the net. Theoretically, such a neural net should be able to solve any instance of the problem, of any length, and be perfectly generalizable, without any need for pre-training. Here, the algorithm is pre-known and the neural net is merely a compiler to implement higher-order functions of the symbolic mind. 

\uline{Problems of pattern recognition whose algorithms ought to exist but cannot be written in symbols}: There might be a class of problems like face recognition, whose exact algorithm or a hand-crafted representation is not known. Here, the machine is simply shown instances of the problem, and learns from it. If a perfect algorithm exists, the machine forms a suitable lattice configuration in its higher-dimensional space that remains equivariant across computational iterations.

\uline{Problems whose algorithms provably don't exist:} There might be a class of problems  whose exact algorithm is not known, and it is also provably impossible to write one. Here, the connectionist theory reduces to a mere approximation and cannot provide an exact solution.

\uline{Problems whose algorithms might exist but we are not so sure:}  A neural net may be able to relax into a perfect solution very quickly for every instance of the problem which is easy to verify. Even if it does, how does one merely look at the weights of the net and mathematically prove that it has indeed learnt the algorithm? This seems difficult because testing a machine on infinite instances of the problem (in our lifetimes) is not the same as proving it has learnt a true universal algorithm. Given two infinite dimensional spaces, each of whose elements have a precise bijection with the other, it is possible to come up with a neural operator. If one can prove that a specific neural operator exists for a particular problem, then there might be some solace\cite{kovachki2023neural}. 

\uline{A canonical view of algorithms:} Men of logic tend to describe algorithms as a set of  rules one `ought to follow' to arrive at a correct answer to the given problem. A proof of an algorithm is thus the mathematical construct that the algorithm is indeed correct no matter what the input. However, mathematics itself is a human construct, and it makes sense to define algorithms in a language of symbols. 

It also does not appear to explain intuition, and self-belief: if an intelligent being is guided by an internal set of principles which it cannot describe in symbols, but merely as a `state of being', how can symbols even describe the actions the being takes?  A higher principle of selflessness, appears to overcome what one's external environment tells him to do. The means to suppress one's natural instincts in favor of internal principles might thus be a desirable property to possess. 

Even if we accept the view that the brain itself runs same sort of symbolic algorithms (as a higher order mental function) to arrive at an identical solution, we cannot explain this phenomenon of intuitive inference. Perhaps, it means that rules are not the only way to get to the correct answer. Perhaps, the brain `glides' in a continuous vector space by the means of intuition and arrives at the answer.

A digital machine does require the laws of boolean algebra to function. However, an analog machine can simply operate on continuous voltages and currents to arrive at an answer.  A learning machine thus built on canonical locks  does not need to follow symbolism: one may arrive at the answer by running the machine for a long time, and then verifying the answer at the output terminals.

There is no way to verify it, unless one can build a geometric proof that each possible symbolic algorithm maps to a lattice configuration, and shows that the net reaches that exact same configuration with a perfect zero error. In this view, one substitutes symbols with lattices of mental imagery. It  also prevents the requirement for feeding the net with a symbolic program in the first place, for the weights of the net themselves are the algorithm. 

This leads one to confront a very uncomfortable question: If families of algorithms map to families of part-whole hierarchies, how does one know the ground truth of the higher dimensional  part-whole vectors (for supervision) in the first place? The answer is that one does not, he merely learns it. One has to abandon the notion of proof. It is still possible to invent a connectionist net to relax to a perfect solution, but one cannot prove it. Perhaps, it is a black box of uncertainty: the moment you poke it, it will give you an answer, but whether that answer is what it actually computed may differ. 

\subsection{Reading the wholes of a part in the hierarchy}

Assuming that the internal representations of the net can form part-whole hierarchies, we will now describe the mechanisms of how to `read' which parts belong to which wholes. 
\begin{figure}[ht]
    \centering   
    \includegraphics[width=0.2\textwidth]{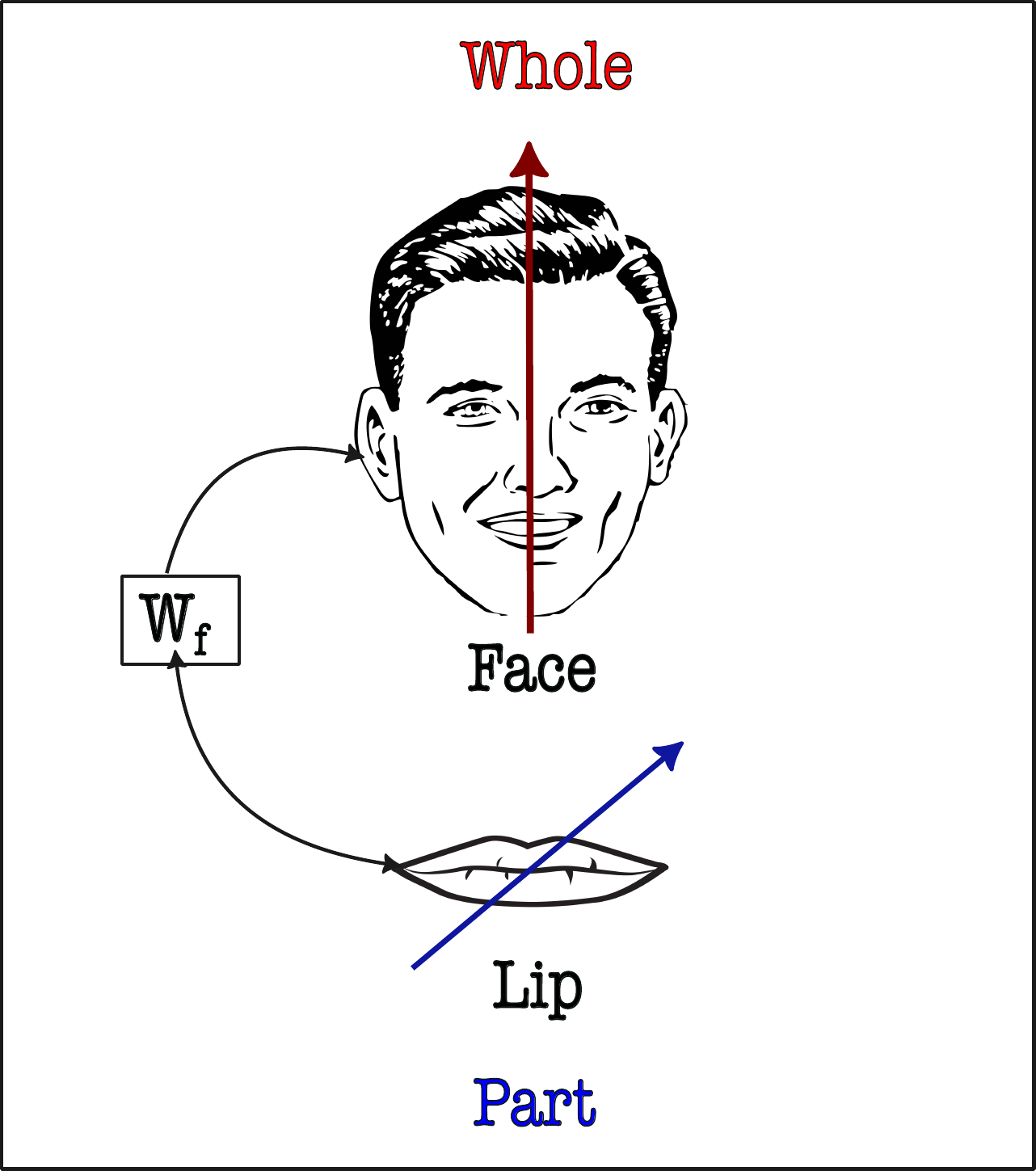} 
    \caption{A single part sending a vote to the higher level whole}
    \vspace{-1em}
    \label{fig:intro_bottom_up}
\end{figure}

Fig\ref{fig:intro_bottom_up} shows a single part vector (lip) getting transformed by bottom-up neural net $W_{f}$ to yield the whole vector (face). An initial idea was to connect all the parts (in lower level) to all the wholes (in higher level). The actual connectivity pattern could be either convolutional or attention, but for the context of this discussion, we shall assume the operation in a single window where the net attends to elements in a scene (for eg, inside a single kernel of a CNN, or the entire global attention operation).

\begin{figure}[ht]
    \centering   
    \includegraphics[width=0.4\textwidth]{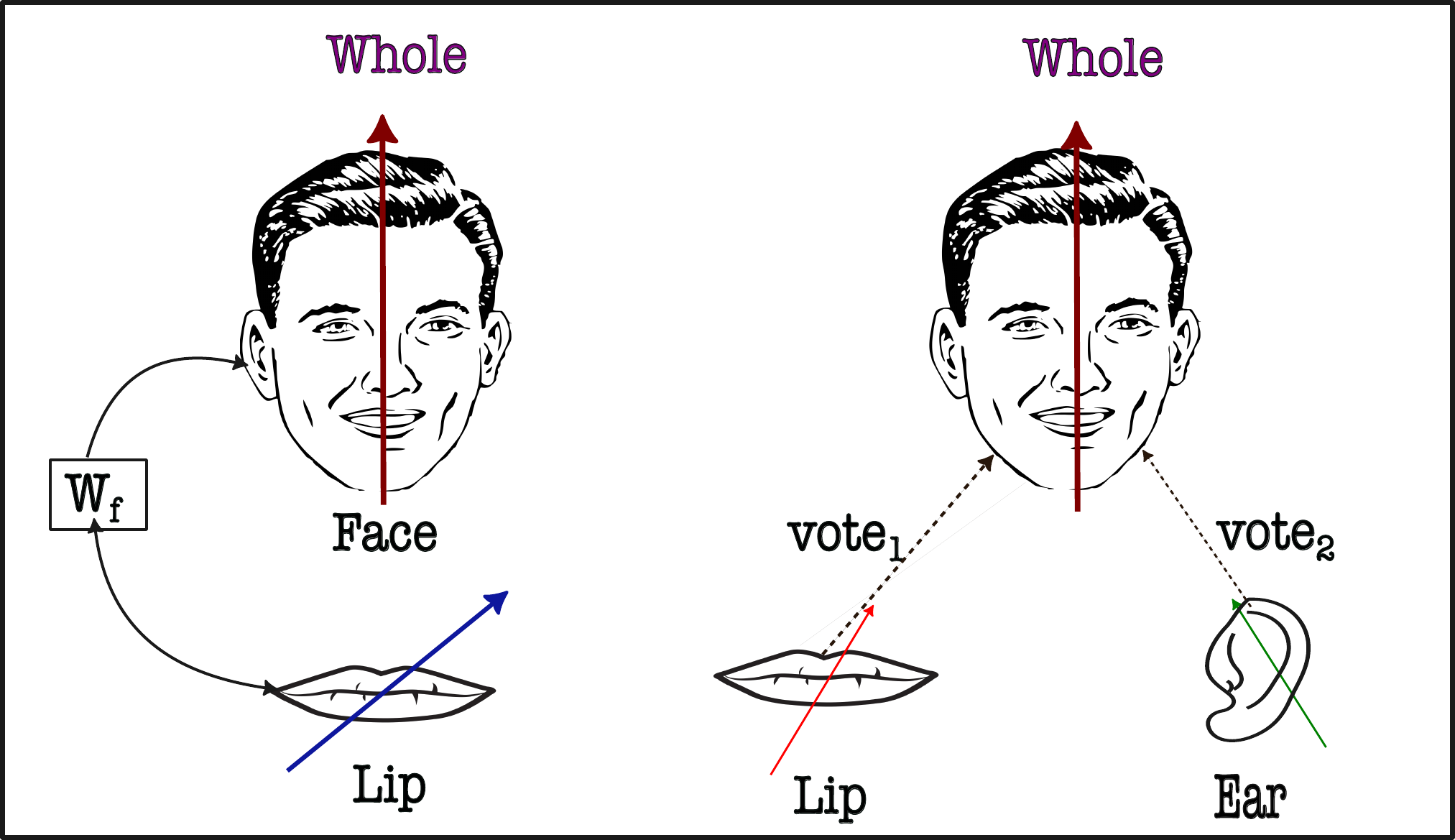} 
    \caption{The buggy capsule-voting mechanism}
    \vspace{-1em}
    \label{fig:capsule_vote}
\end{figure}

Fig\ref{fig:capsule_vote} describes the issues with capsule-style EM/attention based routing. Here, (on the right), we show two separate parts (lip and ear) producing their own votes for the whole (face). If the two parts produced the same whole vector (after transformation), and prediction matched  the original whole vector, then the parts were said to `agree' to that whole. However, that form of representation quickly ran into two troubles (which could not be resolved by training at scale):

% \vspace{-0.2em}
\begin{itemize}[itemsep=0.1em]

\item There was only one whole (and hence one neuron) representing the face. This meant that both lip/ear had to `route' information to the same higher level neuron. Given $N$ parts, $M$ wholes, the routing mechanism needs to encode $N\times M$ decisions, a constraint which just expects too much from  gradient descent. 
\item The transformation matrix $W_f$ was assumed to be $4 \times 4$ dimensional. The decision was made taking inspiration from computer graphics: a 3D object can be rotated/translated from one point to another using a $4 \times 4$ transformation matrix.
\end{itemize}

The notion of a $4 \times 4$ transformation matrix is however deeply flawed. First, it restricts the transformation matrix to be low dimensional, which means that the quality of representation the net learns is sub optimal. Second, even if we assume it to be higher dimensional $d \geq 4$, it quickly becomes infeasible to visualize the $d \times d$ matrix. 

Finally, even if its dimensions are reduced to $4 \times 4$ (using tsne, pca etc), the learnt representation does not ensure that the correct indices in the matrix encode rotation/translation, and that the last row is purely homogeneous. One then requires a clever alignment mechanism (like Umeyama's) to see what the net has actually learnt which just takes too much compute. The idea that the brain does inverse graphics albeit powerful does not mean that the brain is actually doing those transformations in a three dimensional world.

% \vspace{-0.2em}

\begin{figure}[ht]
    \centering   
    \includegraphics[width=0.4\textwidth]{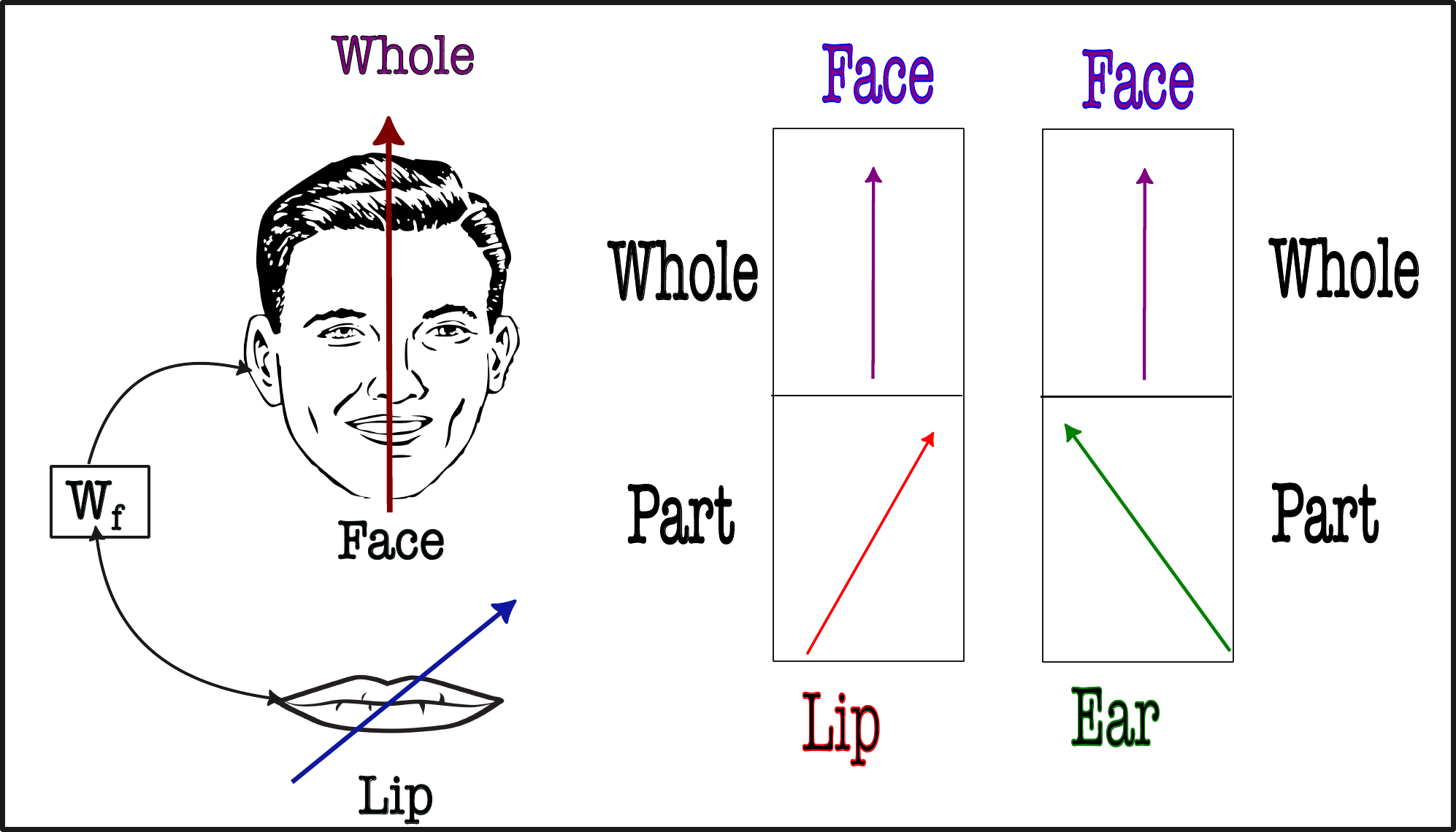} 
    \caption{Eliminating part-to-whole transformation matrices}
    \vspace{-1em}
    \label{fig:glom_vote}
\end{figure}

The key realization is to eliminate the transformation matrix, and instead assume an alternate representation: one that relies on replication of the whole vector across all the parts. Fig\ref{fig:glom_vote} shows one such case: there are now two blue vectors at the level of the whole (face). One can contrast it with the capsule representation in Fig\ref{fig:capsule_vote}, which contained only one vector for the whole (face).

The way one measures whether two parts belong to the same whole has now changed: We can directly compare the two face vectors with themselves, and see if they are close to each other. Assuming a more general case of $M$ whole vectors, one can assign membership to $K$ clusters by using a k-means algorithm. 

The key differences of this form of representation are  :

\vspace{-0.2em}
\begin{itemize}[itemsep=0.1em]

\item There is no more routing. There is a separate whole vector being predicted for every part vector. 
\item The value of K is not known before decoding. Clustering happens after the recognition has happened. There is no need of bounding boxes or Hungarian matching.
\end{itemize}
\vspace{-0.2em}

This is different from slots which require the number of slots to be known a priori, and rely on competitive learning to latch correct slot onto correct object. This results in an issue where a slot latches onto more than one object and needs to be split using some heuristic or learnt rule.Our mechanism  however is gratefully inspired from neural-synchronization\cite{akorn}.

\subsection{Brain as a generative model}
Earlier, Boltzmann machines proposed the idea that a common set of neurons could be used for both perception and generation. For a given input process, a bottom-up process generates recognition  and a top-down generative process provides a feedback loop of neural activity. Similarly, the brain appears to have both feedforward and feedback connections which operate asynchronously.

Taken together, both these systems iterate over several iterations, to reach what one may now term as `islands of agreement'\cite{glom,modi2024asynchronous}. This idea also inspired the invention of Helmholtz Machines. Later, AlexNet popularized the effectiveness of a simple bottom-up approach, because it was simpler to implement and train. The notion of a top-down feedback loop was abandoned\footnote{It was replaced by feeding top layer activations of a layer in a \textit{previous timestep} into  a lower layer at next timestep. However, this requires backpropagation to take a time-out and unfold the computational graph to form a directed acyclic graph.  } and the ideas of a purely feedforward approach (trained at scale) became pre-dominant. 

Our implementation of canonical locks is closer in spirit to   GLOM\cite{glom} (than transformers)  as follows:

\vspace{-0.2em}
\begin{itemize}[itemsep=0.1em]

\item There are now separate bottom-up and top-down neural nets which can synchronize each other. 

\item We have removed the notion of singular fixation: i.e. the assumption that a neural net can look at an input image only once. 
\end{itemize}
\vspace{-0.2em}

Next, we  consider how issues with the bottom-up net (in the original GLOM formulation) were resolved\cite{glom}. This has connections to mean field theory. Later, we   will consider the full architecture of the model.

\subsection{Issues with the mean field approximation}

Let us consider how the part-whole hierarchy may be formed inside the net. 
\begin{figure}[ht]
    \centering   
    \includegraphics[width=0.3\textwidth]{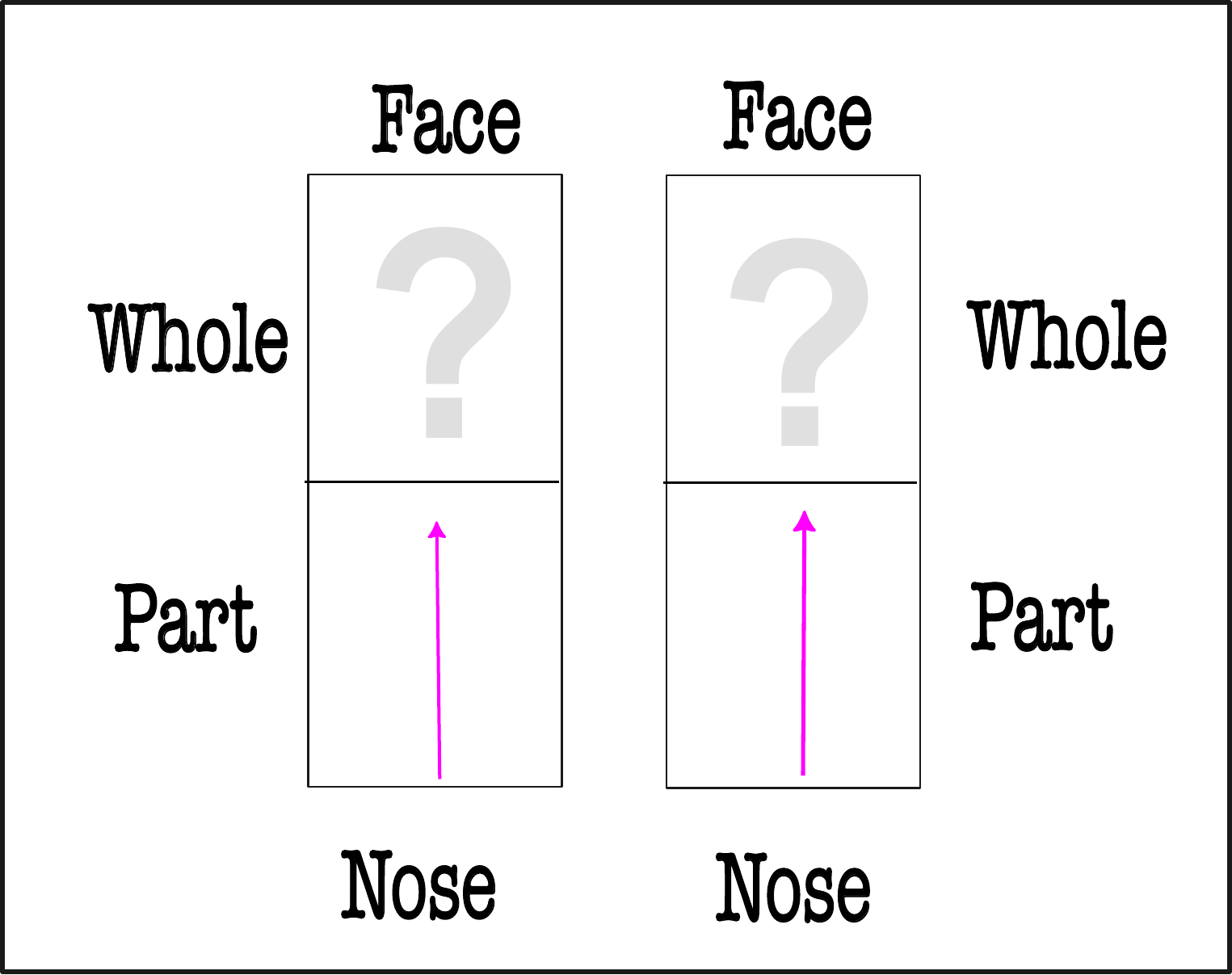} 
    \caption{A two level hierarchy showing formation of whole vectors from part vectors}
    \vspace{-1em}
    \label{fig:bu_hierarchy}
\end{figure}

\begin{figure}[ht]
    \centering   
    \includegraphics[width=0.3\textwidth]{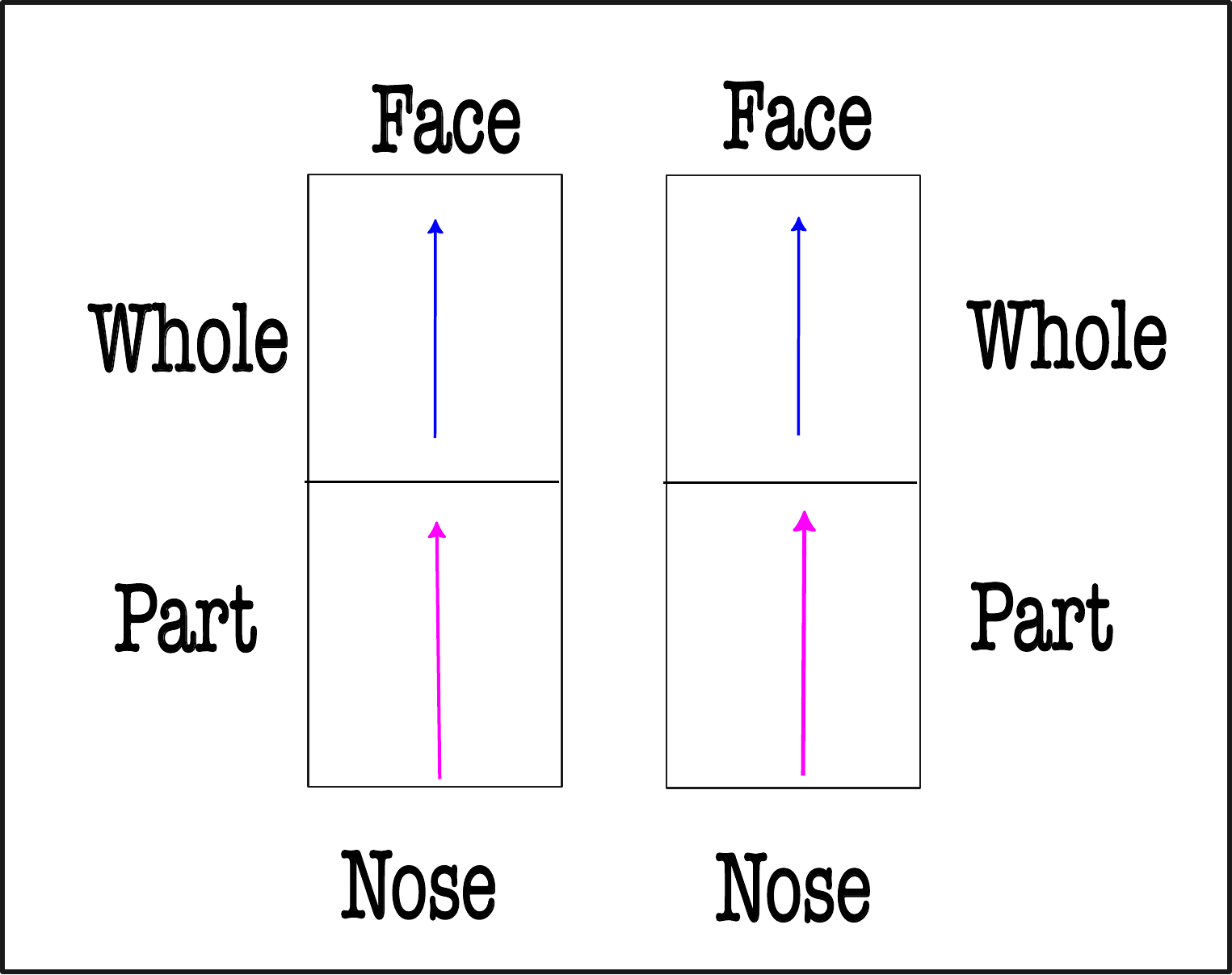} 
    \caption{Representational collapse in mean field approximation}
    \vspace{-1em}
    \label{fig:bu_collapse}
\end{figure}

Fig \ref{fig:bu_hierarchy} shows a two-level hierarchy. The lowest level consists of two pink vectors belonging to a single part (nose). We are concerned with how to decide the higher level vector of the face (marked by a question mark). Mean field approximation suggests that the higher level vector could be computed as the average of the lower level vector. This results in the picture shown in Fig \ref{fig:bu_collapse}. 

Fig \ref{fig:bu_collapse} shows the collapse of the hierarchy due to mean field approximation. This is not desirable since this is the same as \ref{fig:synchrony}, which means that no amount of useful information can be encoded into the hierarchy.

\begin{figure}[ht]
    \centering   
    \includegraphics[width=0.3\textwidth]{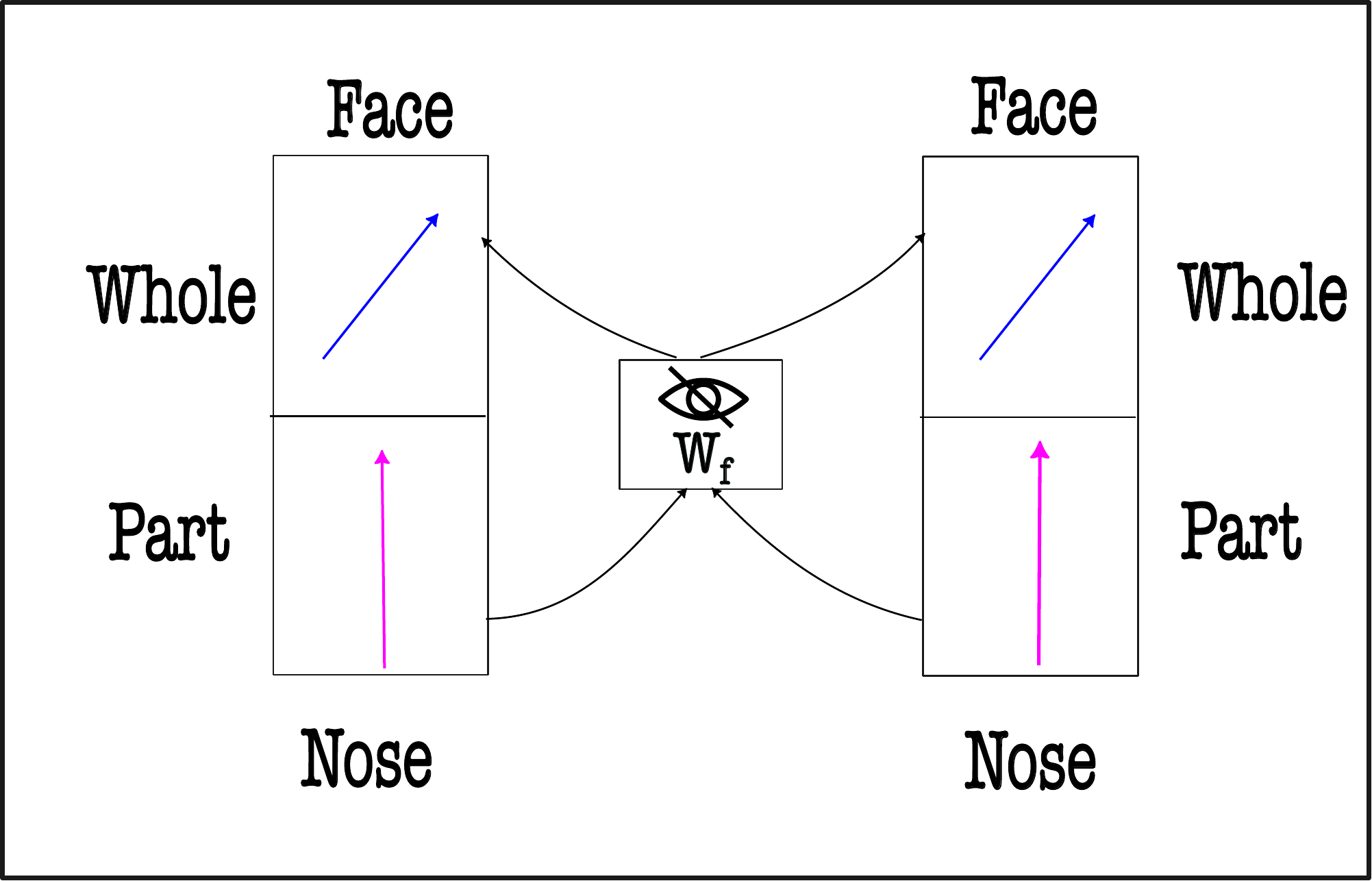} 
    \caption{An alternate solution involves bottom-up net $W_f$}
    \vspace{-1em}
    \label{fig:bu_rotate}
\end{figure}

The solution is simply to `transform' all the lower level part vectors into higher level whole vectors using a bottom-up neural net with weights $W_f$. This results in several properties:

\vspace{-0.2em}
\begin{itemize}[itemsep=0.1em]

\item The bottom up net $W_{f}$ is shared across all part vectors at a particular level. 
\item One must guarantee that $W_f\neq \mathbf{I}$ in order to prevent collapse in Fig \ref{fig:bu_collapse}. 
 
\end{itemize}
\vspace{-0.2em}

This is easy to guarantee if the bottom-up net is a multi-layered neural net with non-linearities and  initialized with random weights (as long as weights of different layers are not identical). Next, we introduce Asynchronous Perception Machines (APM's). The key upgrade over \cite{modi2024asynchronous} is that we have now implemented canonical locks. 

\section{Asynchronous Perception Machines}
\label{sec:apm}
\begin{table*}[t]
\caption{Mathematical notation used in the rest of this paper}
\label{tab:notation}
\vspace{2pt}
\centering
\small
\renewcommand{\arraystretch}{1.25}

\begin{tabularx}{\textwidth}{@{} l >{\raggedright\arraybackslash\hsize=1.15\hsize}X p{0.3cm} l >{\raggedright\arraybackslash\hsize=0.85\hsize}X @{}}
\toprule
\textbf{Symbol} & \textbf{Description} & & \textbf{Symbol} & \textbf{Description} \\
\midrule
$\mathbf{x} \in \mathbb{R}^{H \times W \times C}$ & Input image & & $W_{f}= \{W_{f,1}, \dots, W_{f,L}\}$ & All bottom-up weight matrices \\
$l \in \{1, \dots, L\}$ & Current level in the hierarchy, $L$ total levels & & $W_{b}= \{W_{b,1}, W_{b,2}, \dots, W_{b,L}\}$ & All top-down weight matrices \\
$(i, j,l)$ & Spatial grid coordinates at level $l$ & & $W_{outer}= \{W_{e}, W_{f}\}$ & Outer-loop weights \\
$f_e(\cdot)$ & Encoder & & $W_{inner}= \{W_{b}\}$ & Inner-Loop weights \\
$g_{f,l}(\cdot)$ & Bottom-up Net at level $l$ & & $\text{Opt}_{outer}$ & Optimizer for $W_{outer}$ \\
$W_{f,l}$ & Bottom-up Net's weights & & $\text{Opt}_{b,l}$ & Optimizer for top-down net in level $l$ \\
$h_{f,l}(\cdot)$ & Top-Down net at level $l$ & & $\text{Opt}_{b}$ & Collection of all top-down optimizers \\
$W_{b,l}$ & Top-Down net's weights & & & \\
\bottomrule
\end{tabularx}
\end{table*}

\begin{figure*}[ht!]
    \centering   
    \includegraphics[width=\textwidth]{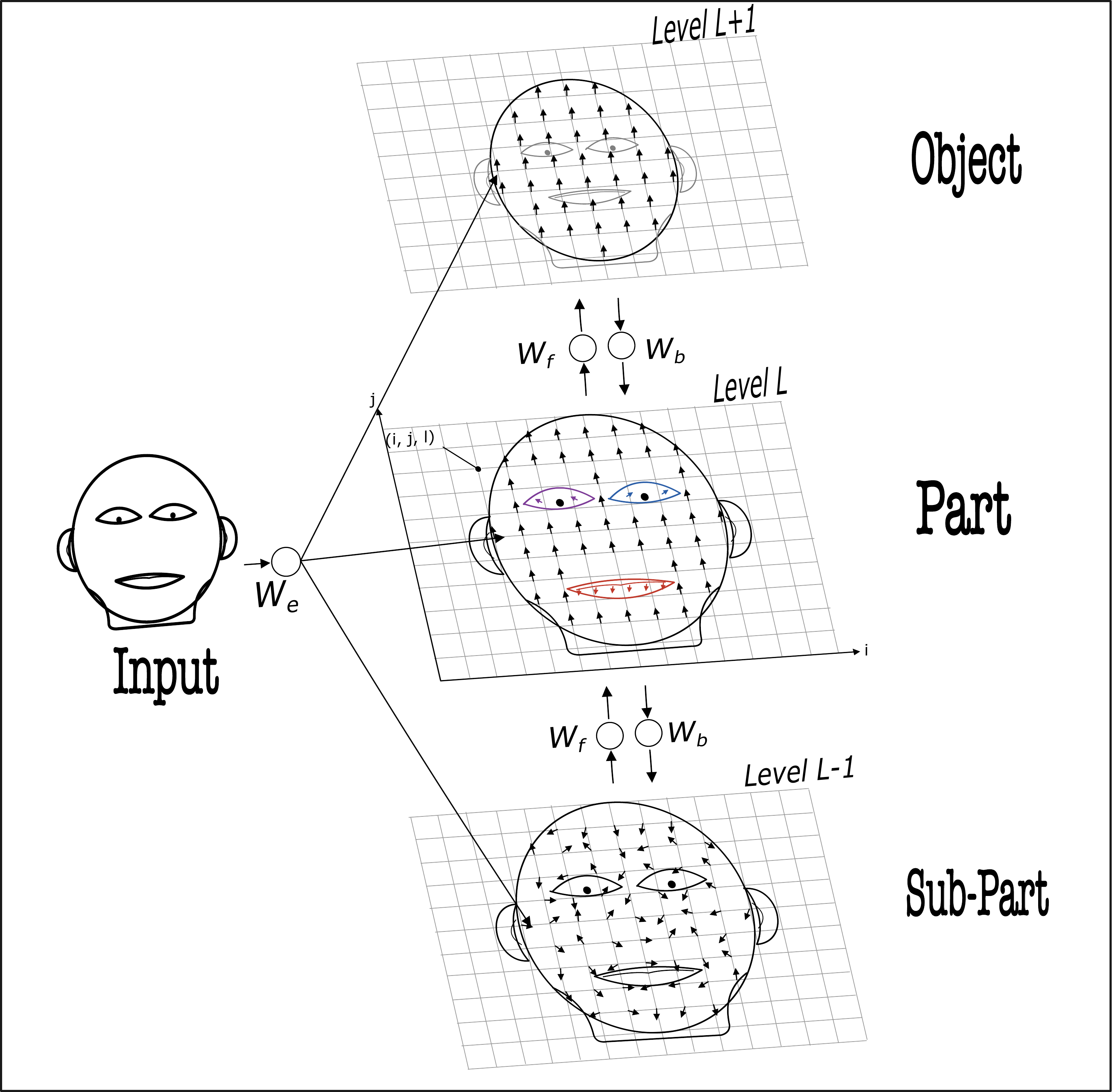}
    \caption{Asynchronous Perception Machine. $W_e$ is encoder, $W_f$ is the bottom-up net and $W_b$ is the top-down net. They may be implemented with either   convolutional or attention based connectivity.}
    \vspace{-1em}
    \label{fig:final_arch}
\end{figure*}

First, we present an overview of APM's architecture in Fig \ref{fig:final_arch}. APM consists of three sets of weights. The input $x$ (eg, image of a face)
 is transformed by an encoder $f_e(\cdot)$ (with weights $W_e$) to produce a set of vectors spanning the part-whole hierarchy. For an image of height $H$, width $W$, and $L$ levels of the hierarchy, there are $H \times W \times L$ such vectors of d dimensions each.  Fig \ref{fig:final_arch} shows three such levels labelled as sub-part, part, and object levels. Note that $W_{e}$ is shared across all these locations.

 At the lowest level $L-1$, all the vectors in the face are arranged randomly, much like needles in a magnetic field. As one goes up to level $L$, the vectors of left eye, right eye, lips begin to point in same directions, indicating that they are part of the same organ. Finally, at the top level $L+1$, all the vectors of the face point in the same direction, indicating that they belong to the same object (called face). A pseudocode for APM's operation has been described in Algorithm \ref{alg:glom}.

\begin{algorithm}[ht!]
\caption{APM Pseudocode}
\label{alg:glom}
\begin{algorithmic}[1]
\REQUIRE Image $\mathbf{x}$, $W_{outer} = \{W_e, W_f\}$, $W_{inner} = \{W_b\}$,
         $\text{Opt}_{outer}$, $\text{Opt}_b$, levels $L$, label $y$

\STATE \textbf{Outer Loop:}

\STATE $\mathbf{z} \leftarrow f_e(\mathbf{x}; W_e)$ \COMMENT{encode image using shared encoder}
\STATE Freeze $W_{inner}$, Unfreeze $W_{outer}$ 
\STATE Predict part-whole hierarchy vectors $v_{i,j,l}$ \COMMENT{for all $(i,j,l)$ locations}

\STATE \textbf{Inner Loop:}
\STATE Freeze $W_{outer}$
\FOR{each pair of levels $(l, l+1)$}
    \FOR{$t = 1$ to $T$}
        \STATE $v_p \leftarrow$ vectors at level $l$
        \STATE $v_w \leftarrow$ vectors at level $l+1$
        \STATE $v_p \leftarrow \text{update}(v_p)$
        % \STATE $v_w \leftarrow \text{update}(v_w)$
        \STATE $\langle v_p, v_w \rangle \leftarrow \text{agreement}(v_p, v_w)$
        \STATE $v_w \leftarrow \text{update}(v_w)$
    \ENDFOR
    \STATE Update $W_{inner}$ using $\text{Opt}_b$
\ENDFOR

\STATE $v_L \leftarrow$ vectors at level $L$
\STATE $\hat{y} \leftarrow \text{Decoder}(v_L)$
\STATE $\mathcal{L}_{outer} \leftarrow \text{Loss}(\hat{y}, y)$
\STATE Update $W_{outer}$ using $\text{Opt}_{outer}$

\end{algorithmic}
\end{algorithm}
% \vspace{-3em}

 Successive levels of the hierarchy are connected by a bottom-up net $g_{f,l}(\cdot)$ (with weights $W_{f,l}$) and a top-down net $h_{f,l}(\cdot)$ (with weights $W_{b,l}$). The bottom-up net transforms the part vectors into whole vectors, whereas the top-down net transforms the whole vectors into part vectors. Together, these two nets undergo a settling process to achieve canonical locks. This process is repeated for all pairs of levels $(l, l+1)$ in the hierarchy (line 6 of Algorithm \ref{alg:glom}). We term this process as the `agreement procedure', and describe it in detail in a later section. Finally, the vectors at the topmost level $L$ are fed to a downstream decoder to produce a task-specific prediction $\hat{y}$ (line 14 of Algorithm \ref{alg:glom}).
 
 \subsection{The Weight Update rule }

APM is optimized in two loops: an inner loop and an outer loop. The inner loop is the agreement procedure which is run for $T$ iterations to achieve canonical locks. The only weights updated in this loop are the top-down weights $W_{b,l}$. Each top-down net contains its separate optimizer $\text{Opt}_{b,l}$. Note that this loop does not require any task-specific labels.  

The outer loop consists of two sets of weights: (i) the encoder weights $W_e$ and (ii) the bottom-up weights $W_f$ (for every level $l$). Both $W_{e}/W_{f}$  are optimized using a single optimizer $\text{Opt}_{outer}$. Note that the outer loop requires task-specific labels to compute the loss $\mathcal{L}_{outer}$, which in turn is used to update $W_{e}/W_{f}$. One may also perform a self supervised task like masked image reconstruction. Optimization can be done using a learning algorithm (like backpropagation). 

Next, we discuss  the agreement procedure.

\subsection{The agreement procedure}
\label{sec:agreement}
\begin{figure}[ht!]
    \centering   
    \includegraphics[width=0.49\textwidth]{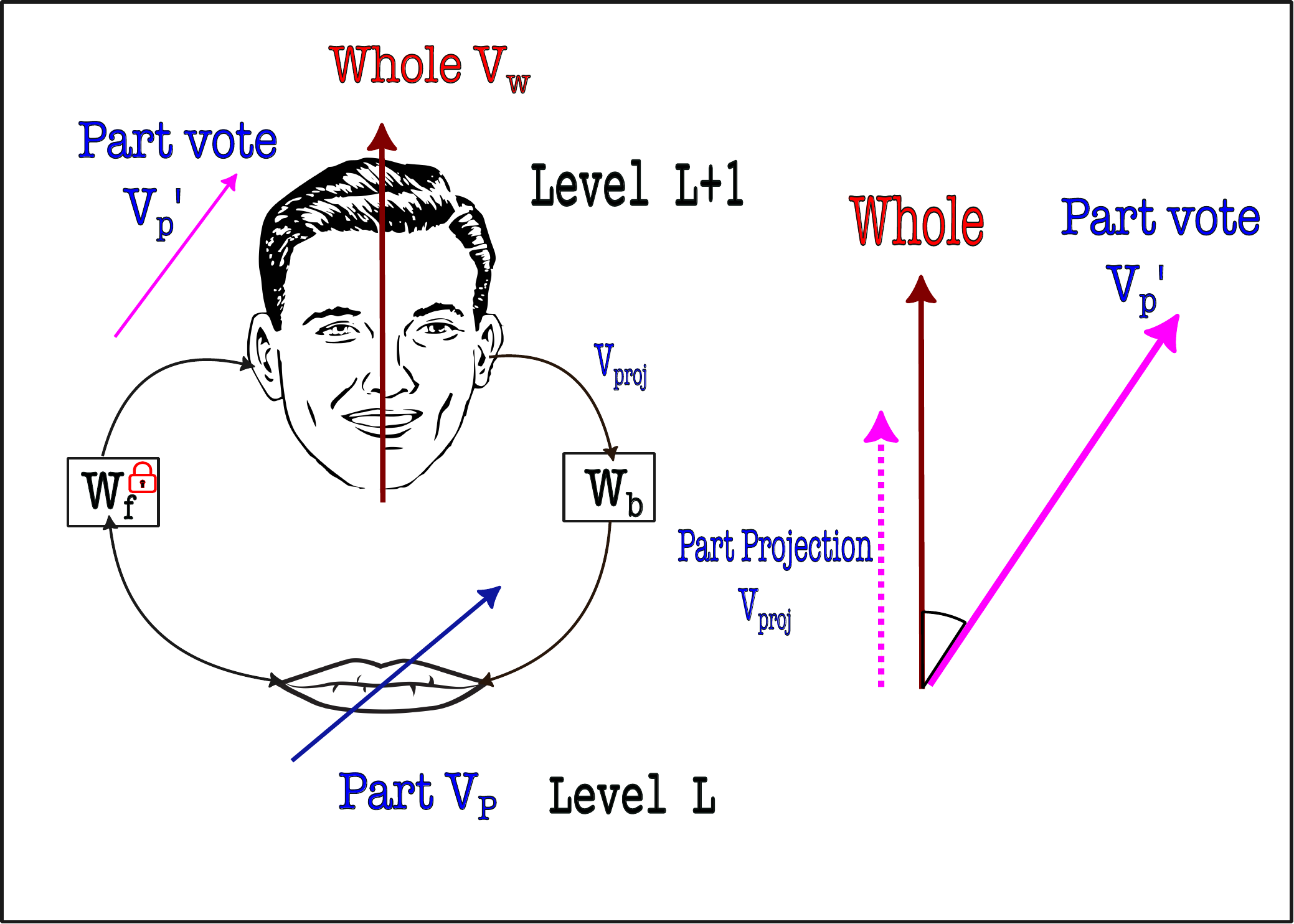} 
    \caption{The agreement procedure.}
    \vspace{-1em}
    \label{fig:agreement}
\end{figure}

Fig\ref{fig:agreement} shows the agreement procedure. We consider a part vector $v_{p}$ sampled from level $l$ and a whole vector $v_{w}$ sampled at level $l+1$. First, $v_p$ self-attends to other part vectors in a circular  neighborhood of radius $r$\footnote{r is set as a hyperparameter, or locations of vectors $v_{p}$ attends to could themselves be learned}. A similar operation is performed for $V_{w}$. Mathematically, we denote this by:

\vspace{-1em}
\begin{subequations}
\begin{align}
v_p &\leftarrow \sum_{q \,:\, \lVert q-p \rVert \le r} \mathrm{softmax}_q\big(\mathrm{sim}(v_p, v_q)\big)\, v_q
\label{eq:part_aggregation}
\end{align}
\end{subequations}

% v_w &\leftarrow \sum_{q \,:\, \lVert q-w \rVert \le r} \mathrm{softmax}_q\big(\mathrm{sim}(v_w, v_q)\big)\, v_q

% v_w &\leftarrow \sum_{q \,:\, \lVert q-w \rVert \le r} \mathrm{softmax}_q\big(\mathrm{sim}(v_w, v_q)\big)\, v_q
The above equation only enables communication within the `same' level \textit{l}. Please note that this neighborhood information aggregation could (in principle) be implemented by  self-attention\cite{Vaswani2017AttentionIA}, Kuramoto synchronization or any other similar mechanism.

Next, we consider the communication agreement `between' different levels \textit{l} and \textit{l+1}. This communication happens as the information is piped through both bottom-up and top-down nets. Note that during this procedure the bottom-up weights $W_f$ are frozen, and only the top-down weights $W_b$ are updated.

First, the part vector $v_{p}$ 
produces a part vote 
$v_{p}'$ as:

\vspace{-1.5em}
\begin{equation}
v_p' = v_p \, W_f
\end{equation}

The agreement between $v_p'$ and $v_w$ is measured by computing the projection of the part vector onto the whole vector as:

\vspace{-1em}
\begin{equation}
v_{proj} = \frac{v_{part} \cdot v_{whole}}{\lVert v_{whole} \rVert^2} \, v_{whole}
\end{equation}
\vspace{-1em}

A perfect agreement is said to be achieved when $v_{proj} = v_{w}$. The whole vector $v_{w}$ is itself updated as a combination of the part vote $v_{proj}$ and interactions with other whole vectors in the (radial) neighborhood. The update rule is given by:

\begin{subequations}
\begin{align}
v_w &\leftarrow \sum_{q \,:\, \lVert q-w \rVert \le r} \mathrm{softmax}_q\big(\mathrm{sim}(v_w, v_q)\big)\, v_q \\
v_w &\leftarrow \left\langle v_w + v_{proj} \right\rangle
\end{align}
\label{eq:whole_update}
\end{subequations}
\vspace{-2em}

The addition operation between $v_{w}$ and $v_{proj}$ induces a form of coupling between levels $l$ and $l+1$. Indeed, there might be many ways to combine (and separate) information from two sources like $v_{w}$ and $v_{proj}$ together. But, the set transformer paper\cite{lee2019set} proved that a simple addition operation is sufficient to achieve universal approximation. It also bears deep mathematical connections to Arnold Kolmogorov's superposition theorem (as acknowledged in the original GLOM paper\cite{glom}), which in turn inspires this choice. An interpretation can also be found in the appendix.

Next, $v_{w}$ is fed back through the top-down net $W_{b}$ to produce a new part vector $v_{part-out}'$ as:
\begin{equation}
v_{part-out}' = v_{w} \, W_b
\end{equation}

We estimate the regression loss between the original part vector $v_{part}$ and the new part vector $v_{part-out}'$ as:
\begin{equation}
\mathcal{L}_{agreement} = \lVert v_{part-out}' - v_{part} \rVert^2
\end{equation}

and the top-down weights $W_b$ are updated using backpropagation. Next, we \textit{detach} the part vector $v_{part-out}'$ from the compute graph and use it to refine the part vector $v_{part}$ itself. This refinement procedure is called \textit{the slurp update},  which we explain next.

\subsection{The Slurp Update}
\label{sec:slurp_update}

Given an input part vector $v_{p}$, and the output of the top-down net $v_{part-out}$, we want to `nudge' the part vector $v_{p}$ `slightly' in the direction of the $v_{part-out}$. It is very tempting to make one rapid update to the part vector $v_{p}$ by just setting $v_{p} = v_{part-out}$. However, this is not ideal, since then the top-down net gets zero loss, and cannot be updated beyond one iteration. Instead, we propose the slurp-update.

\begin{figure}[ht!]
    \centering   
    \includegraphics[width=0.4\textwidth]{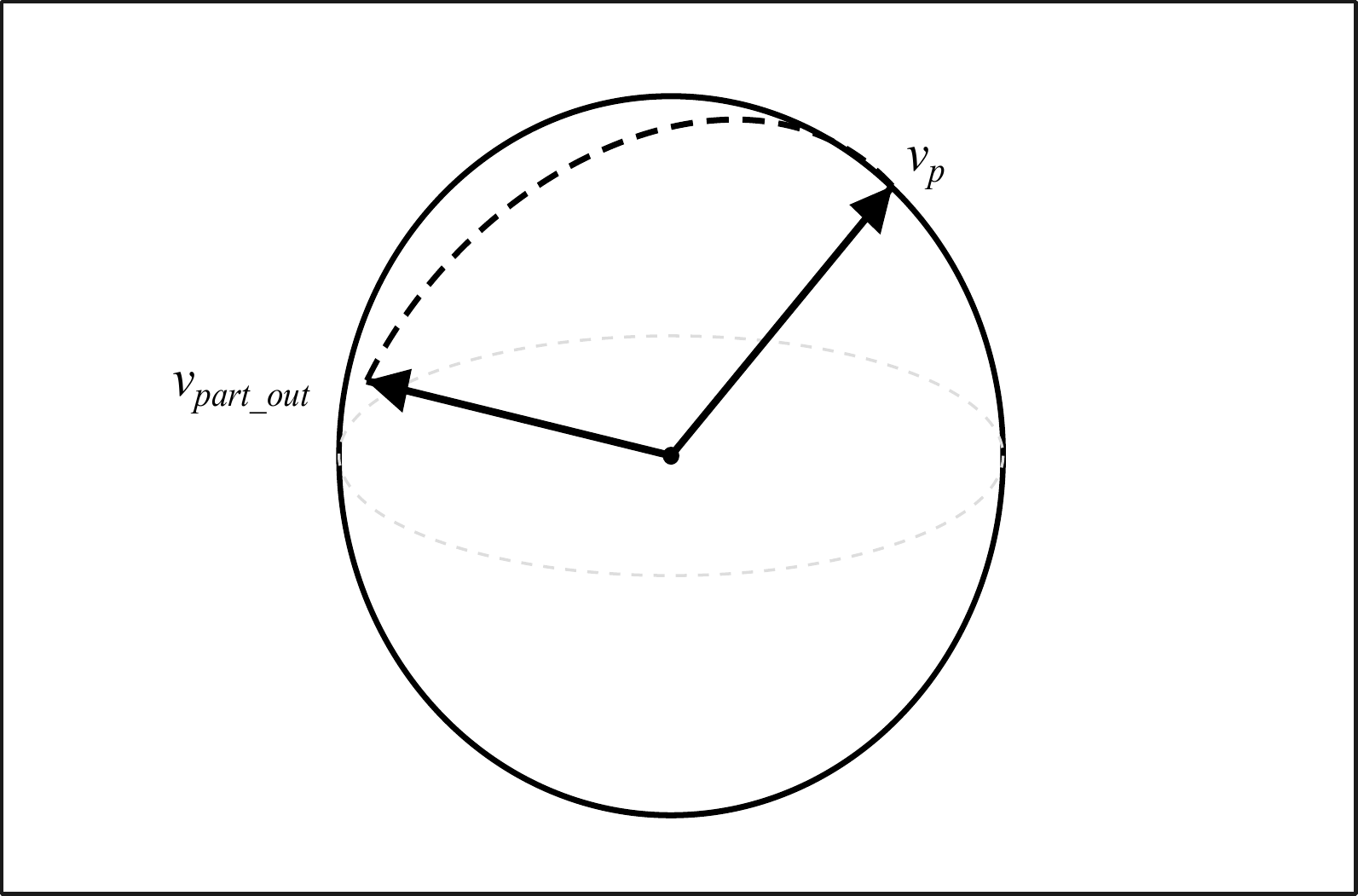} 
    \caption{The slurp update.}
    % \vspace{-1em}
    \label{fig:slurp_update}
\end{figure}

First, we normalize both $v_{p}$ and $v_{part-out}$ so that they lie on surface of a unit hypersphere. Next, we draw an arc on the sphere between $v_{p}$ and $v_{part-out}$. The slurp update is to move the part vector `slightly' along this arc towards $v_{part-out}$. The length of the arc can be said to be $L$. We can divide this into small $n$ segments of length $l$, such that $n \times l$  = $L$. The vector $v_{p}$ is stepped towards $v_{part-out}$ by $l$ length on the geodesic. The number $n$ (a hyperparameter) controls how many partitions of the arc are made.

Thus, the information recirculates freely across the bottom up and top down nets, followed by the slurp update. This is repeated for $T$ iterations, and does not need any labels. Finally, in the outer loop, the weights $W_{e}$ and $W_f$ are themselves updated via backpropagation.

Note that backpropagation updates level $l+1$ with larger gradients than level $l$. Gradient normalization/clipping is an option, but it is not enough to bypass backpropagation's reliance on chain rule: successive multiplications of gradients across layers result in weaker gradient signal. This means that the signal from the bottom-up net is weaker than the signal from the top-down net. To compensate, one thus needs to run the equation \ref{eq:part_aggregation} for a greater number of iterations than equation \ref{eq:whole_update}.

One issue with $L_{agreement}$ is that it only uses the available information in part vectors. There is no constraint on the architectures of the bottom-up and top-down nets  which ensures that  agreement happens in `least' number of iterations. To resolve this, one needs to consider how the architecture of bottom-up and top-down net should be designed. This leads us to the matter of `when might perfect agreement arise?'

\subsection{Promoting Perfect Agreement}
\label{sec:perfect_agreement}

\begin{figure}[ht]
    \centering   
    \includegraphics[width=0.2\textwidth]{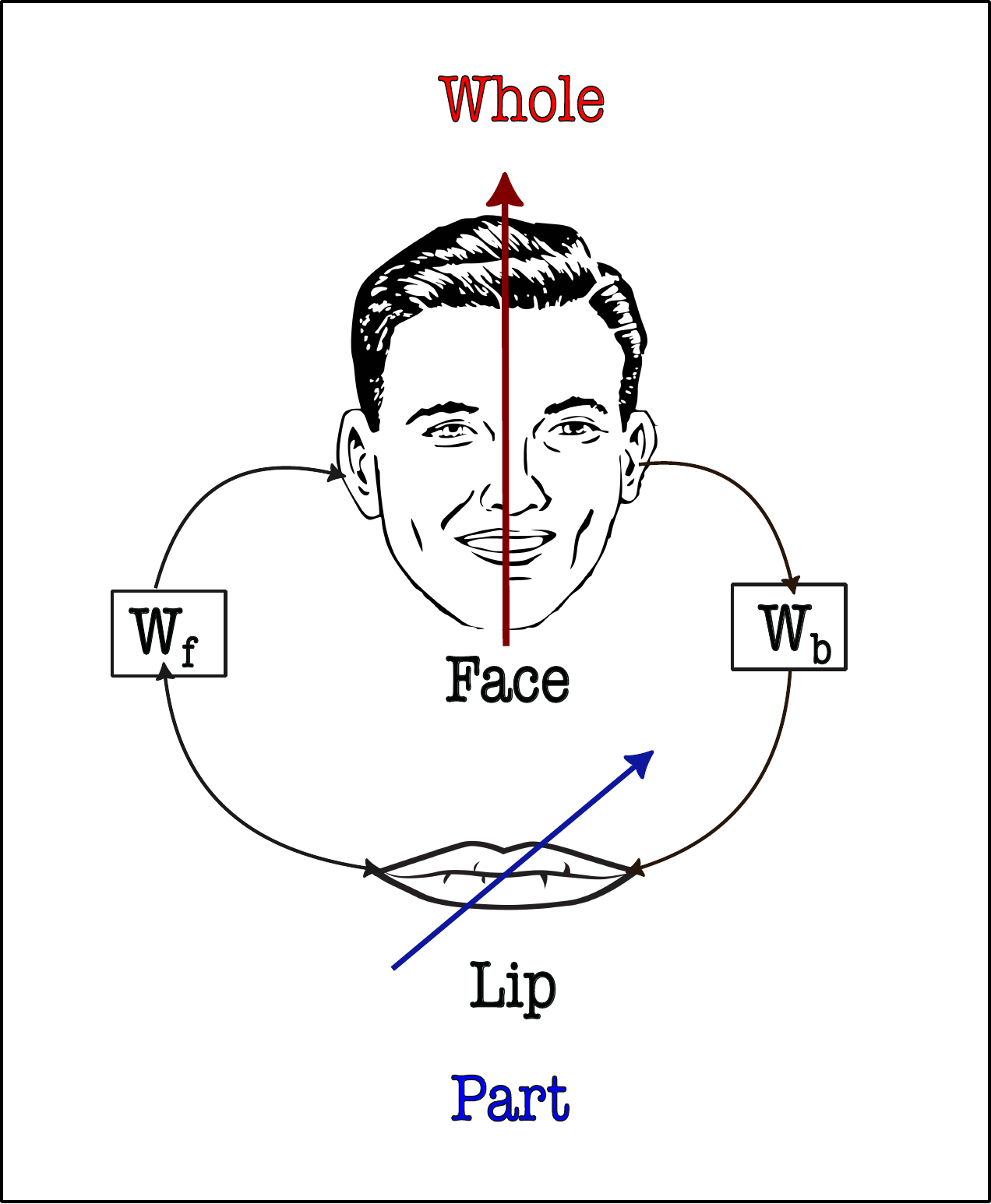}
    \caption{An analysis of the case of perfect agreement.}
    \vspace{-1em}
    \label{fig:bu_fixed}
\end{figure}

We consider the simplest case of a single part vector $v_p$ and a single whole vector $v_w$. In the case of perfect agreement, the following conditions shall hold:

% \vspace{-2em}
\begin{subequations}
\begin{align}
v_w = W_f v_p\\
v_p = W_b v_w 
\label{eq:perfect_agreement_equations}
\end{align}
\end{subequations}

The regularization constraint to prevent collapse is that $W_f \neq \mathbf{I}$ and $W_b \neq \mathbf{I}$.  Substituting $v_p$, we obtain $v_w = W_f W_b v_w$. This implies that $W_f W_b = \mathbf{I}$, which means that $W_b = W_f^{-1}$. In principle, one could enforce this by estimating the inverse of $W_f$ in each iteration of the inner loop, and regressing $W_b$ towards that. However, this does not work due to two reasons:

\vspace{-0.2em}
\begin{itemize}[itemsep=0.1em]

\item The bottom-up net consists of multiple layers with non-linearities. There is no guarantee that during every iteration, an inverse actually exists.
\item Even if it did, taking an inverse of a matrix is computationally expensive, which means that it does not parallelize as well on GPUs as compared to matrix multiplications.

\end{itemize}
\vspace{-0.2em}

\begin{figure}[ht!]
    \centering   
    \includegraphics[width=0.5\textwidth]{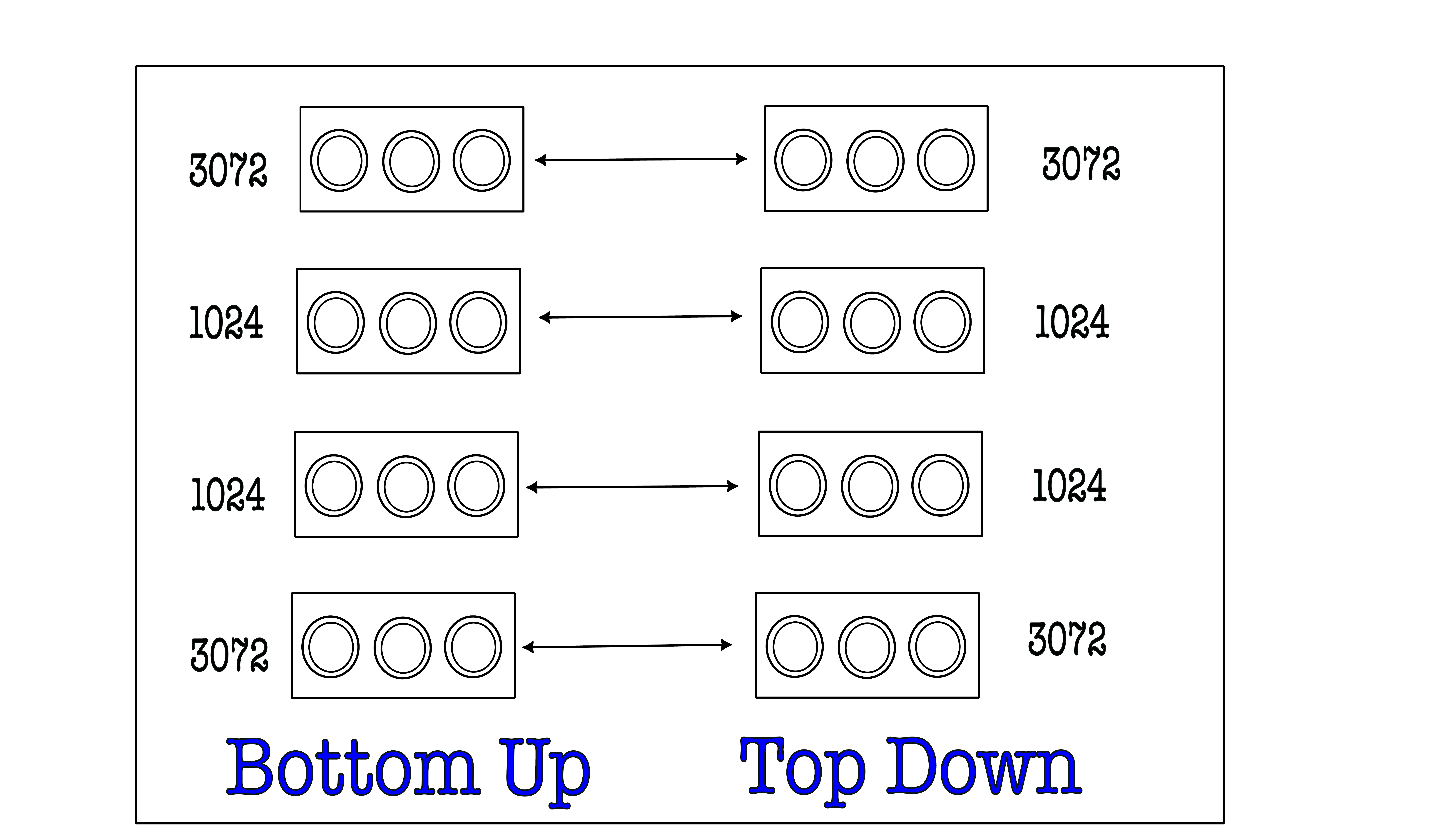} 
    \caption{Combining  bottom-up and top-down nets together. They are identical in structure. Each box shows the number of  neurons present in that layer. Each neuron is marked as a `double circle' to indicate that it may be used to `both' present an input to the net, as well read output from it. Intermediate activation functions are omitted for clarity.}
    \vspace{-1em}
    \label{fig:mirror_symmetry}
\end{figure}

One tempting way out of this hurdle is to enforce the condition $W_f W_b = \mathbf{I}$ on each layer of the nets. We term this procedure as Deep Weight Alignment. However, for this constraint to work, the bottom-up and top-down nets must be layer-wise symmetric as shown in Fig\ref{fig:mirror_symmetry}\footnote{We tried to use the same net as a top-down and bottom-up net, but it did not work as well as keeping them separate. }.

During each iteration of the inner loop, we enforce:

\vspace{-2em}
\begin{subequations}
\begin{align}
W_{pred} = W_f W_b \\
L_{weight}= \lVert W_{pred} - \mathbf{I} \rVert^2
\label{eq:deep_weight_alignment}
\end{align}
\end{subequations}

The total loss $L_{inner}$ is then given by:
\begin{equation}
L_{inner} = L_{agreement} + \lambda L_{weight}
\end{equation}
% \vspace{-1.5em}

\begin{figure*}[ht!]
    \centering
    % Scaling to 0.48 of textwidth keeps it perfectly inside one 
    % column
    % \vspace{-1em}
    \includegraphics[width=\textwidth]{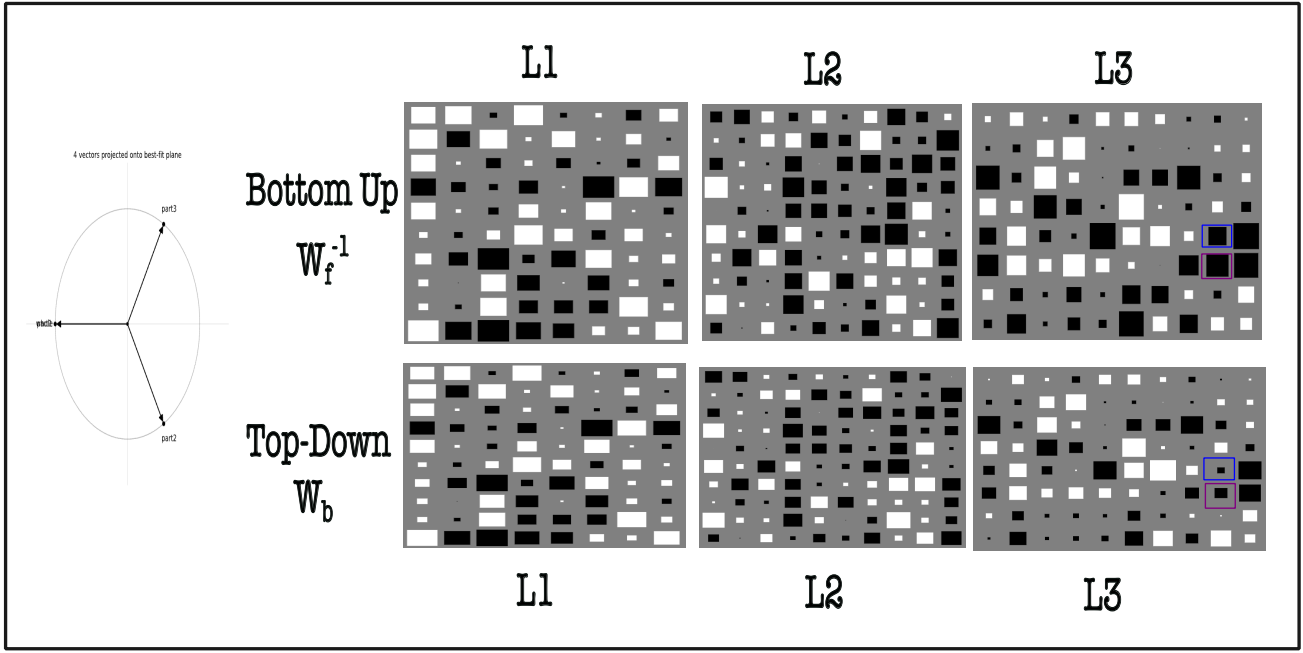} 
    % \vspace{-1em}
    \caption{Hinton Diagrams for synchronization: The top-down and bottom-up net  can drive each other to identical weights. They follow signed symmetry: the magnitude of sign of the weight in each box is the same. However, the magnitude of the weight may be different. For eg, the thickness of blue highlighted square, is different in layer L3 of both nets. }
    \vspace{-1em}
    \label{fig:sync_plot}
\end{figure*}

Note that neither $L_{agreement}$ nor $L_{weight}$ require any labels. The inner loop is thus a self-supervised procedure which promotes islands of agreement inside the net\cite{modi2024asynchronous, glom}.

Connections to Random Feedback Alignment: On a first glance, the  Deep Weight Alignment procedure appears similar to the idea of random feedback alignment\cite{lillicrap2016random}. However, there are several differences: 

\begin{itemize}
    \item The top-down net is a mirror symmetric copy of the bottom-up net. 
    \item The top-down net carries agreement signal, and not the gradient signal. 
    \item The bottom up net is not trained at all. It is merely a pathway for the lower level part vectors to send neural activity to the  higher level whole vectors.
\end{itemize}

We acknowledge that the resultant system is biologically implausible. First, it creates the illusion that brain contains two pathways which must be symmetric.  But such observations have not been made in neuroscience. Second, the notion that a part of the circuit (bottom up net) is never trained seems absurd. 

However,  this does appear to work better as compared to training both the nets together.  The job is far from done, and there must be some other  alternative mechanisms by which brain does credit-assignment that does  not rely on gradient descent. A weight update rule $W = W - \eta \nabla_W L$ means that both the weights and gradients are stored, which in turn implies there must be two weights on each  synapse\footnote{This is different from slow-fast weights, here one weight is actual knowledge, other is the `change' required to modulate it.}. This seems even more absurd. A better alternative might be to let the weight matrix `itself know' how to update itself, without ever estimating the gradients\cite{irie2022modern}. This is not a path we have yet explored.

Connections to Kuramoto Oscillatory Neurons: The agreement procedure communicated in Section \ref{sec:agreement} is similar to the idea of Artificial Kuramoto oscillatory neurons\cite{akorn}. However, the differences are :

\begin{itemize}
    \item Synchronization happens between two different levels of the hierarchy, and not within the same level.
    \item A top-down net is responsible for the synchronization. Therefore, the application of minimum description length principle to $v_{proj}$ is not enforced directly, but is an implicit consequence of enforcing $L_{agreement}$.
    \item The constraint that the part vote needs to be in same direction as the whole vector $v_w$ need not hold: the net can decide the angle (aka canonical lock) between the two vectors during gradient descent.
\end{itemize}

\subsection{Hinton diagrams for synchronization}

Hinton diagrams were a connectionist weapon designed  during the times of the symbolist-connectionist war\cite{fodor1988connectionism}. A Hinton diagram is a 2D grid of squares, visualizing the weights of the net. For an N by N weight matrix, the total grid is of size N by N. Each element of the grid is a square. The thickness of the square denotes the magnitude of the weight, and the color (black or white) denotes the sign of the weight. Multiple such squares can then be plotted for each layer of the bottom up/ top-down  net. We show one such plot in Fig\ref{fig:sync_plot}.

We perform the following experiment: (i) the part vectors in level $l$ and level $l+1$ are initialized randomly. (ii) Both bottom-up and top-down nets are initialized randomly. The weights of the bottom-up net are frozen. (iii) The agreement procedure  is performed and the weights of both the nets (After T steps) are plotted. $L_{inner}$ is used to update the weights of the top-down net. 

Fig\ref{fig:sync_plot} shows the two rows with weights $W_f^{-1}$ and $W_b$ visualized. One can see that the weights of top-down net $W_b$ are driven to the weights of $W_f^{-1}$, which is the inverse of the bottom-up net. The only differences are in the layer $L_{3}$ where the thickness (hence the magnitude) of the weights is different.

However, the agreement procedure seems to promote signed symmetry between the two nets, i.e. the `color' of the boxes comes out to be identical. We tested our procedure with different initializations of $W_f$ prior to the agreement procedure. Similarly, we tested running the optimization for more than a million iterations, and the results appear somewhat consistent. This suggests that the agreement procedure can indeed allow bottom-up net and top-down net to drive each other to synchrony.

If the vectors at part level and whole level are entirely random, then the agreement procedure still converges to their mean. This means that merely examining the loss is not enough to reveal the quality of islands being formed. However, it the data possesses some structure, (for eg, images), we can indeed see islands of agreement to emerge\cite{modi2024asynchronous, glom}. 

A way to promote formation of islands is to enforce two constraints: (i) attention must be restricted to a circular local neighborhood, much akin to  restricted Boltzmann machines (ii) The radius of the neighborhood is gradually increased as one goes up the different layers of $W_e$. This is similar to the idea of receptive fields in convolutional nets. The key issue with CNNs was that they traded off the resolution of the input in favor of increasing channel count in the later layers. This led to a loss of spatial information, which had to be compensated for  via skip connections from a layer $l$ of encoder to  layer $l$ of decoder. 

Our mechanism is different in two ways: (i) As information flows through $W_{e}$ there is no loss of spatial resolution (ii) a standard softmax looks at all the locations, and weighs the incoming signal accordingly. This is flawed because it guides the agreement towards a single mean and leads to collapse. 

Our procedure instead relies on resonance\cite{liu2026krause}\footnote{A principle that in turn is inspired from echo chambers.}: a part vector looks at all its neighbors, and chooses the `most dominant one'. Once it has done this selection, it does not care about what all other part-vectors are saying. Our part aligns itself in the direction of the dominant neighbor. 

In the initial layers of $W_e$, there are several part vectors around a given part-vector, pointing in randomly different directions. Thus, there is no structure in the internal representation. Thus, the attention mechanism needs to seek consensus from k top neighbors. As we go up in the hierarchy, the circular neighborhoods we tap into gradually increase in radius. The value of k is gradually stepped down during training from 4 to 1 because by then the system has already beaten the natural entropy, and cluster formation has already happened.

\subsection{Alternating signed symmetry and amplification circuits}

Looking closely in Fig\ref{fig:sync_plot} shows that the weights in different layers appear to alternate their signs. For eg, the top left square in $W_f^{-1}$ (L1) is white, whereas it alternates to black in $W_f^{-1}$ (L2), and white again in $W_f^{-1}$ (L3). It is also interesting to see that the strength of the weights (the thickness) of the square in L1 is much stronger than $L3$. Similarly, the weights of top-down net $W_b$ are of lower magnitude in L2/L3 throughout. This leads to several observations:

\begin{itemize}
    \item Top-down net was trained with backpropagation with the first layer having the strongest gradients. As we go deeper, the gradients become weaker and the ability of the net to have signed bit flips in the weights get reduced which in turn hurts performance. Top-Down net is naturally `weaker' than the bottom-up net. 
    \item An electromagnetic wave travels in a medium with alternative positive and negative flips. So perhaps, weights can be thought of as a medium and the top-down/bottom-up net as electromagnetic waves. 
    \item A combination of top-down and bottom-up net then creates a standing wave, which is a form of resonance. If both waves support each other, one can overcome the weakness of the top-down net. If not, then the top-down net is drowned out by the bottom-up net.

\end{itemize}

\begin{figure*}[ht!]
    \centering
    \setlength{\fboxsep}{12pt}
    \setlength{\fboxrule}{0.8pt}
    \fbox{%
        \includegraphics[width=0.9\textwidth]{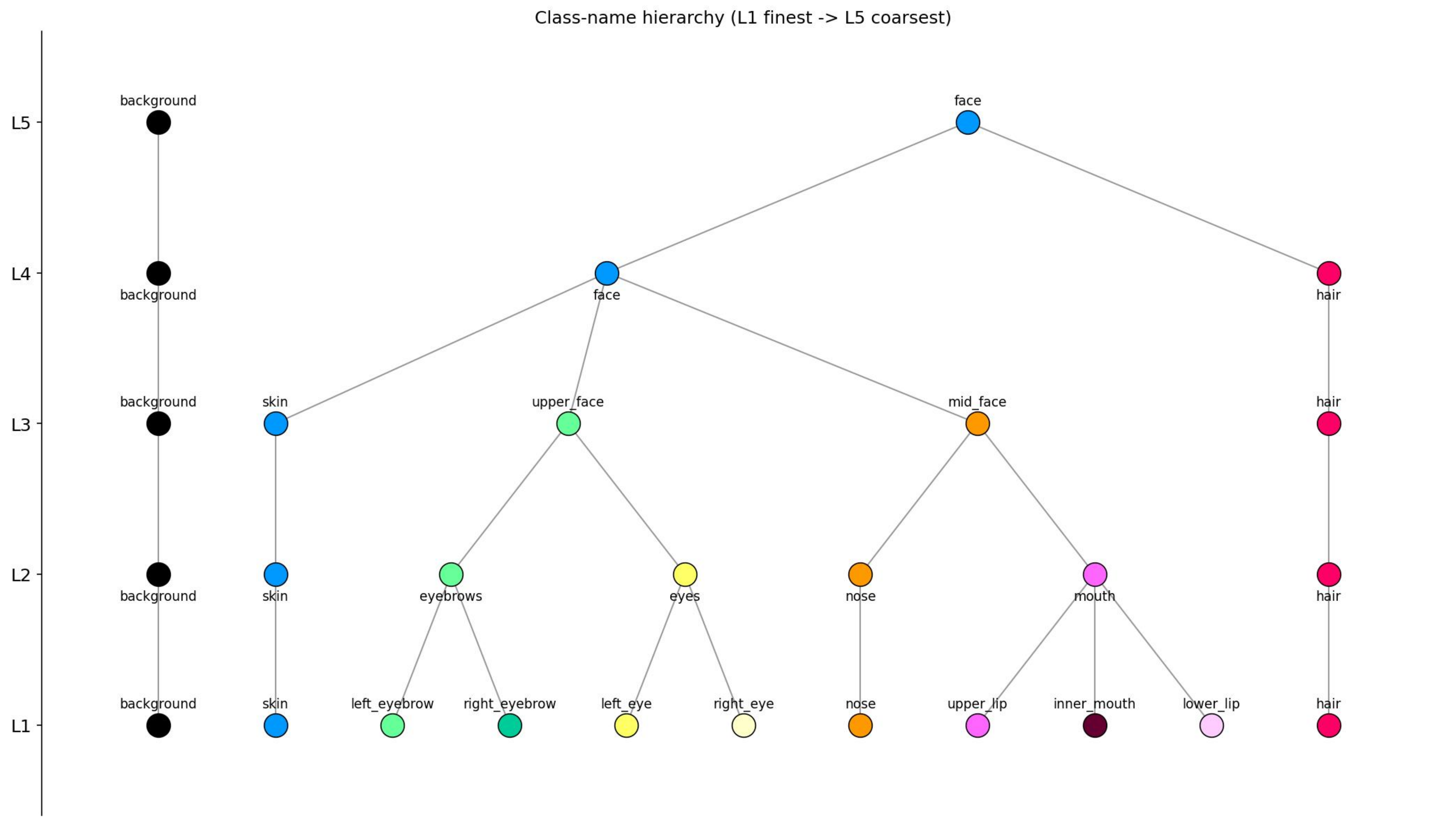}%
    }
    \caption{The fixed class-name hierarchy (L1 finest to L5 coarsest) that the model's 5 classification heads are trained against, shown as a dendrogram.}
    \label{fig:class_hierarchy_dendrogram}
\end{figure*}

    A careful tuning of the wavelengths of travelling waves in both nets is thus required to make sure that the peaks coincide with the place where output is read off the neural circuit. Similarly, a standing wave forms alternative crest and troughs. It is then possible to `tap into' all the crests of the standing wave to make sure that the decoded representation is the strongest one. A tap at the trough however results in destructive interference which severely diminished the quality of the decoded representation.

The fact that top-down and bottom-up nets create their own travelling waves whose combination results in a standing wave, leads to the question: how do we prevent the crests of the standing wave from blowing up? This is a very important question, since if the crests blow up, then the agreement procedure will not converge. The answer was to simply clamp the weights to a spherical higher dimensional manifold. An alternative is to use Riemann geometry. 

During the agreement procedure, there was another finding. Since the magnitude of the weights of bottom up net were more than the top-down net, it was better to  measure the internal Lyapunov energy of the bottom-up net to make a sense of whether the agreement was converging or not. To compensate for the weaker top-down nets, we simply `amplified' the weights of the top-down net by a fixed factor. Alternatively, one can switch directions from which the backpropagation is performed: first operate on the left side of the Fig\ref{fig:sync_plot} and then on the right side. On average, it results in weights of top-down net to mirror the weights of bottom-up net. 

Let us now assume that weight of bottom up net is $W_{f,i,j,l}$ and weight of top-down net is $W_{b,i,j,l}$. Here, $l$ represents the $l^{th}$ layers of the nets and $(i,j)$ represent the spatial coordinates $(i,j)$ of the weight matrix. The correction factor $\gamma_{i,j,l}$ is given by:

\begin{subequations}
\label{eq:impedance_matching}
\begin{align}
    \gamma_{i,j,l} &= \frac{\lVert W_{f,i,j,l} \rVert}{\lVert W_{b,i,j,l} \rVert + \epsilon} \label{eq:correction_factor} \\
    \widetilde{W}_{b,i,j,l} &= \gamma_{i,j,l} \cdot W_{b,i,j,l} \label{eq:weight_rescaling}
\end{align}
\end{subequations}

The universal approximation theorem leads one to believe that it is possible to abstract away all the families of neural architectures as an abstraction function $f$. However, our experience suggests that the rate at which representations learnt by different architectures converge depends on the nature of the wiring pattern itself.

\subsection{Replication of top-down and bottom-up nets}

Kindly refer again to the architecture in Fig\ref{fig:final_arch}.  Consider the top-down and bottom-up nets between level $l$ and $l+1$ again. One can observe that both the nets are shared across all the locations $(i,j)$ in the spatial grid. We wish to consider various ways in which this weight sharing might occur. 

Firstly, one could replicate top-down and bottom-up net across locations $(i,j)$ in the spatial grid. On a single GPU, this weight copying means that one quickly runs out of memory. However, one gains the ability to exchange information between different copies of the net by weight sharing, which is a much higher bandwidth way of exchanging information than knowledge distillation\cite{glom,hinton2022forward}

Secondly, a single copy of the top-down and bottom-up net is kept. Instead, the location based queries $(i,j)$ are fed as a batched input to the net\cite{modi2024asynchronous}. This saves memory on a single device. 

Finally, among multiple GPU devices, only one copy of top-down and bottom-up net is kept. Inside the device, the parallelism is achieved by feeding different $(i,j)$ queries to the net. Thus, one in principle could reap the benefits of both weight sharing and memory efficiency.

\section{The visual hierarchical parsing}
\label{sec:visual_hierarchy}

In NLP, it is easy to parse sentences into parse trees, since it is possible to define which words form a noun, verb etc. In case of vision, the hierarchies are not so well defined. Similarly, existing datasets only contain up to 3 levels of hierarchy, whereas GLOM explicitly requires 5 levels of the hierarchy.

Towards this end, we consider the CelebA-HQ dataset\cite{karras2017progressive} of faces, and define a 5-level part-whole hierarchy of a face. An example of the hierarchy is shown in Fig\ref{fig:class_hierarchy_dendrogram}. The hierarchy is defined at lowest level  as \emph{ground}, \emph{skin}, \emph{left-eyebrow}, \emph{right-eyebrow}, \emph{left-eye}, \emph{right-eye}, \emph{nose}, \emph{upper-lip}, \emph{inner-mouth}, \emph{lower-lip} and \emph{hair}. The remaining levels are derived by merging the parts into wholes. The hierarchy is as follows:

\begin{itemize}
    \item \textbf{L1} ($11$ classes) --- the raw parser classes above.
    \item \textbf{L2} ($7$) --- the two eyebrows merge into \emph{eyebrows}, the two eyes into \emph{eyes}, and the three lip/mouth classes into \emph{mouth}.
    \item \textbf{L3} ($5$) --- \emph{eyebrows} and \emph{eyes} merge into \emph{upper face}; \emph{nose} and \emph{mouth} merge into \emph{mid face}.
    \item \textbf{L4} ($3$) --- \emph{skin}, \emph{upper face} and \emph{mid face} merge into a single \emph{face}.
    \item \textbf{L5} ($2$) --- \emph{hair} is absorbed into \emph{face}, leaving only \emph{face} and \emph{background}.
\end{itemize}

The instance of the problem presented to the net  might change: given any image of a face, the net must parse it correctly. The individual sizes of clusters (eg, face, nose, etc), as well as their locations may differ.

Fig\ref{fig:celeba_hierarchy_preds} shows the predicted part-whole hierarchy of a face from the APM. Note that all the levels sit above each other. A body part (i,j) at level $L$ knows what sub-part it breaks into by examining the representation (i,j) at level $L-1$. The visual parsing in this representation can then occur in a backward fashion: starting from level $L$, one can examine the representation at level $L-1$ to see what sub-parts it breaks into, and so on until the finest level is reached. The net generalizes to any unseen image, and is not restricted to the training set.

\uline{Backward Parsing:} For a pixel (i,j) at level L, we can know all the children in the parse tree by looking recursively at representation of (i,j) at level L-1, and so on until the finest level is reached. The procedure is repeated for all the locations $(1,1) \leq (i,j) \leq (H,W)$ in the spatial grid. 

We would like to distinguish this form of visual parsing  from the parsing of sentences in \cite{Vinyals2014GrammarAA}. In the latter, one partitions a sentence into words spanning indices (i,j), builds a representation, and runs a linear classifier. Running the net for all pairs gives a score. An assignment problem is then solved to find the best parse tree which follows a correct syntax. Along the similar lines, a similar hungarian assignment procedure is used in object detectors to allocate boxes to ground truth. 

The parsing of images here does not require any assignment, but instead emerges as a consequence of the agreement procedure. The very act of `gluing together'\footnote{Thus folding is the operator of gluing together multiple representations, whereas unfolding is the operator for separating them back into individual pieces. This is inspired from how gluons in physics bind multiple quarks together.} $\langle$sub part, part, object vectors$\rangle$ in a single column guarantees that the parent-child relationship can be read off directly, which should significantly speed up the parsing process. 

\uline{Connections to autoregression}: A seq2seq model relies on linearizing the parse-tree and predicting each token via autoregression. This is slow because tokens in the far future cannot be predicted until tokens  in the past have been predicted. At each timestep, the decoded neural activity is fed back through the network input, which means that the network must be run again. 

The form of hierarchical parsing proposed here is different: the entire parse-tree across all levels is predicted in a single forward pass and refined over multiple iterations of the bottom up and top down agreement process.

\section{Conclusion}

In this paper, we have presented canonical locks as a geometric primitive for learning machines. This allows them to encode part-whole hierarchies in a neural net. Our key result is that it is possible to train a neural net with just one sample (without any sort of initialization from a pretrained set of weights), and generalize to a given test set of images with competitive accuracies. This generalization also extends to rotational groups that increment in multiples of 90 degrees. 

Our running hypothesis is that neural nets are geometrical engines that encode part-whole hierarchies. Overfitting on a single sample leads  to a form of compositional-based learning which learns to combine parts into wholes dynamically\cite{hinton1990mapping, schmidhuber1990towards, hinton2019representing, glom}. Constraining the weights properly reduces the subspace gradient descent has to search, which in turn significantly speeds up learning.

Given that compute is finally cheap, but the solid-state memory is not, it makes sense to explore learning machines which are higher bandwidth, but lower parameterized. It might be possible for APM/s to meta learn  multiple levels of the  part whole hierarchy. They should meta learn their own architecture, learning algorithms, optimizers, and pretraining task as well. Last but not the least, they should meta learn from the meta learner himself\cite{schmidhuber1987evolutionary}.

\bibliography{example_paper}
\bibliographystyle{icml2026}

\newpage

\appendix
\section{Decoding the part-whole hierarchy}
\label{app:decode_snippet}

Let $\hat{y}_l(i,j) \in \{0, \dots, C_l - 1\}$ be the class the APM predicts at pixel $(i,j)$ and level $l$, with $C_1, \dots, C_5 = 11, 7, 5, 3, 2$ the class counts of Sec\ref{sec:visual_hierarchy}. Every level carries its own name table $N_l$, and decoding a pixel is the lookup
\begin{equation*}
(i,j) \mapsto \big( N_1[\hat{y}_1(i,j)],\, \dots,\, N_5[\hat{y}_5(i,j)] \big),
\end{equation*}
whose five entries are the part-whole path of that pixel: the part it occupies at the finest level, and each whole which contains that part as one moves up. Nothing is post-processed to obtain this path. The features are not clustered, the prediction is not matched against a tree, and no assignment is solved. The path is read straight off the five classifier outputs. It is possible to align an open world text-model to each of the locations of the part-whole hierarchies, which means that multiple hierarchies could be decoded together. The decoding scheme shows that the net has the ability to predict the correct vector at the correct location $(i,j,l)$.

The tables $N_l$ must be reported with the predictions, because every level numbers its classes from zero and an id on its own therefore fixes nothing. The dump below gives them for the face of Fig\ref{fig:celeba_hierarchy_preds}. Its \texttt{class\_ids} field is $N_1$ through $N_5$; its \texttt{level\_accs} field records the accuracy of this sample at each level, rising from $0.919$ at L1 to $0.982$ at L5; and its \texttt{hierarchy} field gives one pixel per class present in the face, each row listing the pair $[\,$id, name$\,]$ at all five levels.

The rows show three behaviors worth separating. A pixel may change both its id and its meaning as the level rises: pixel $(88, 89)$ is id $2$ at L1, \emph{left\_eyebrow}, and id $2$ again at L2, but the table underneath has changed and the two eyebrows are now one class, \emph{eyebrows}; the pixel then becomes \emph{upper\_face} at L3 and \emph{face} at L4 and L5. It is a part and a whole at the same time, and which of the two one sees is set by the level one reads. A pixel may equally keep its meaning for several levels: hair stays \emph{hair} up to L4 and joins \emph{face} only at L5. A class may also go unused, since \emph{inner\_mouth}, id $8$ at L1, has no row at all in this face.

An example of a hierarchy decoded using our visual semantic parser has been provided below.

\begin{strip}
{\footnotesize\setlength{\baselineskip}{9.6pt}
\begin{verbatim}
{
  "sample_idx": 221,
  "level_accs": {"l1": 0.919, "l2": 0.900, "l3": 0.916, "l4": 0.969, "l5": 0.982},
  "class_ids": {
    "l1": {"0": "background", "1": "skin", "2": "left_eyebrow", "3": "right_eyebrow",
           "4": "left_eye", "5": "right_eye", "6": "nose", "7": "upper_lip",
           "8": "inner_mouth", "9": "lower_lip", "10": "hair"},
    "l2": {"0": "background", "1": "skin", "2": "eyebrows", "3": "eyes", "4": "nose",
           "5": "mouth", "6": "hair"},
    "l3": {"0": "background", "1": "skin", "2": "upper_face", "3": "mid_face",
           "4": "hair"},
    "l4": {"0": "background", "1": "face", "2": "hair"},
    "l5": {"0": "background", "1": "face"}
  },
  "hierarchy": [
    {"pixel": [209, 112], "l1": [0, "background"],
     "l2": [0, "background"], "l3": [0, "background"],
     "l4": [0, "background"], "l5": [0, "background"]},
    {"pixel": [130, 161], "l1": [1, "skin"],
     "l2": [1, "skin"], "l3": [1, "skin"], "l4": [1, "face"], "l5": [1, "face"]},
    {"pixel": [88, 89], "l1": [2, "left_eyebrow"],
     "l2": [2, "eyebrows"], "l3": [2, "upper_face"],
     "l4": [1, "face"], "l5": [1, "face"]},
    {"pixel": [88, 146], "l1": [3, "right_eyebrow"],
     "l2": [2, "eyebrows"], "l3": [2, "upper_face"],
     "l4": [1, "face"], "l5": [1, "face"]},
    {"pixel": [103, 66], "l1": [4, "left_eye"],
     "l2": [3, "eyes"], "l3": [2, "upper_face"], "l4": [1, "face"], "l5": [1, "face"]},
    {"pixel": [103, 141], "l1": [5, "right_eye"],
     "l2": [3, "eyes"], "l3": [2, "upper_face"], "l4": [1, "face"], "l5": [1, "face"]},
    {"pixel": [124, 98], "l1": [6, "nose"],
     "l2": [4, "nose"], "l3": [3, "mid_face"], "l4": [1, "face"], "l5": [1, "face"]},
    {"pixel": [153, 103], "l1": [7, "upper_lip"],
     "l2": [5, "mouth"], "l3": [3, "mid_face"], "l4": [1, "face"], "l5": [1, "face"]},
    {"pixel": [167, 129], "l1": [9, "lower_lip"],
     "l2": [5, "mouth"], "l3": [3, "mid_face"], "l4": [1, "face"], "l5": [1, "face"]},
    {"pixel": [85, 13], "l1": [10, "hair"],
     "l2": [6, "hair"], "l3": [4, "hair"], "l4": [2, "hair"], "l5": [1, "face"]}
  ]
}
\end{verbatim}
}
\end{strip}

% \clearpage

\begin{strip}
    \centering
    \setlength{\fboxsep}{12pt}
    \setlength{\fboxrule}{0.8pt}
    \fbox{%
        \includegraphics[width=0.6\textwidth]{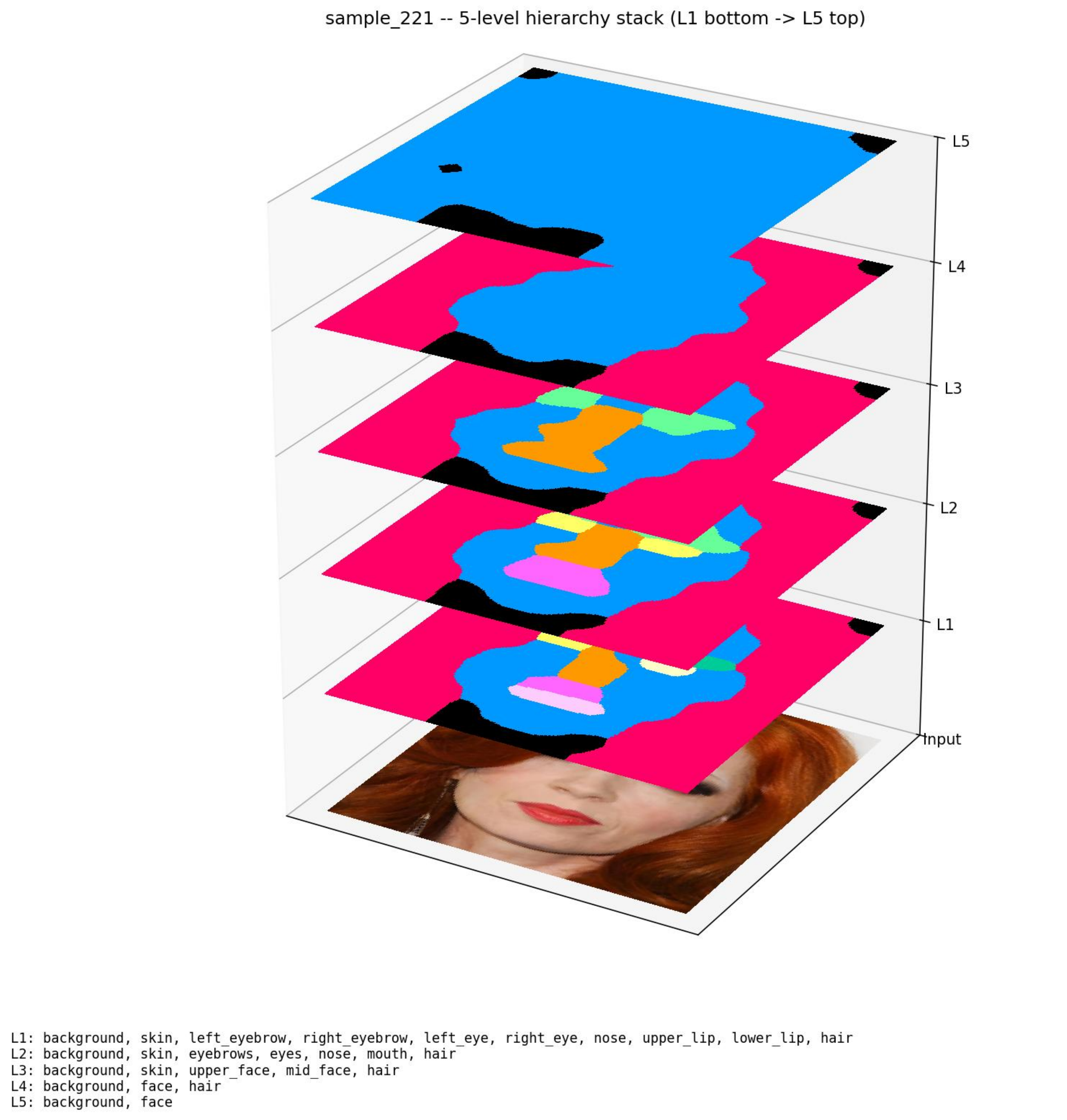}%
    }
    \captionof{figure}{The 5 predicted level maps stacked along a z-axis (L1 finest at bottom, L5 coarsest at top), making the part-whole merging across levels visually explicit: individual parts (eyes, nose, mouth) at the bottom converge into a single face region by the top.}
    \label{fig:stacked_levels_3d}
\end{strip}

\end{document}